\documentclass[10pt,twocolumn,letterpaper]{article}

\usepackage[pagenumbers]{cvpr} %

\usepackage{graphicx}
\usepackage{amsmath}
\usepackage{amssymb}
\usepackage{booktabs}

\usepackage{float,nonfloat}
\usepackage{enumitem}
\newcommand{\secref}[1]{\S\ref{#1}}
\usepackage{multirow}

\usepackage[normalem]{ulem}

\usepackage[pagebackref,breaklinks,colorlinks]{hyperref}

\usepackage[capitalize]{cleveref}
\crefname{section}{Sec.}{Secs.}
\Crefname{section}{Section}{Sections}
\Crefname{table}{Table}{Tables}
\crefname{table}{Tab.}{Tabs.}

\def\cvprPaperID{3276\vspace{-3mm}} %
\def\confName{CVPR}
\def\confYear{2023}

\begin{document}

\title{\vspace{-2mm}GlassesGAN: Eyewear Personalization using Synthetic Appearance Discovery and Targeted Subspace Modeling\vspace{-4mm}}

\author{Richard Plesh\\ %
Clarkson University, USA\\
{\tt\small pleshro@clarkson.edu}
\and
Peter Peer, Vitomir Štruc\\
University of Ljubljana, Slovenia\\
{\tt\small peter.peer@fri.uni-lj.si,vitomir.struc@fe.uni-lj.si\vspace{75mm}}
}

\maketitle

\thispagestyle{empty}
\noindent
\begin{minipage}[b]{\textwidth}
\vspace{-95mm} %
    \begin{minipage}[b]{1.0\textwidth}
    \centering
    \vspace{-30mm}
        \includegraphics[width=0.94\textwidth, trim = 0 0.1mm 0 0, clip]{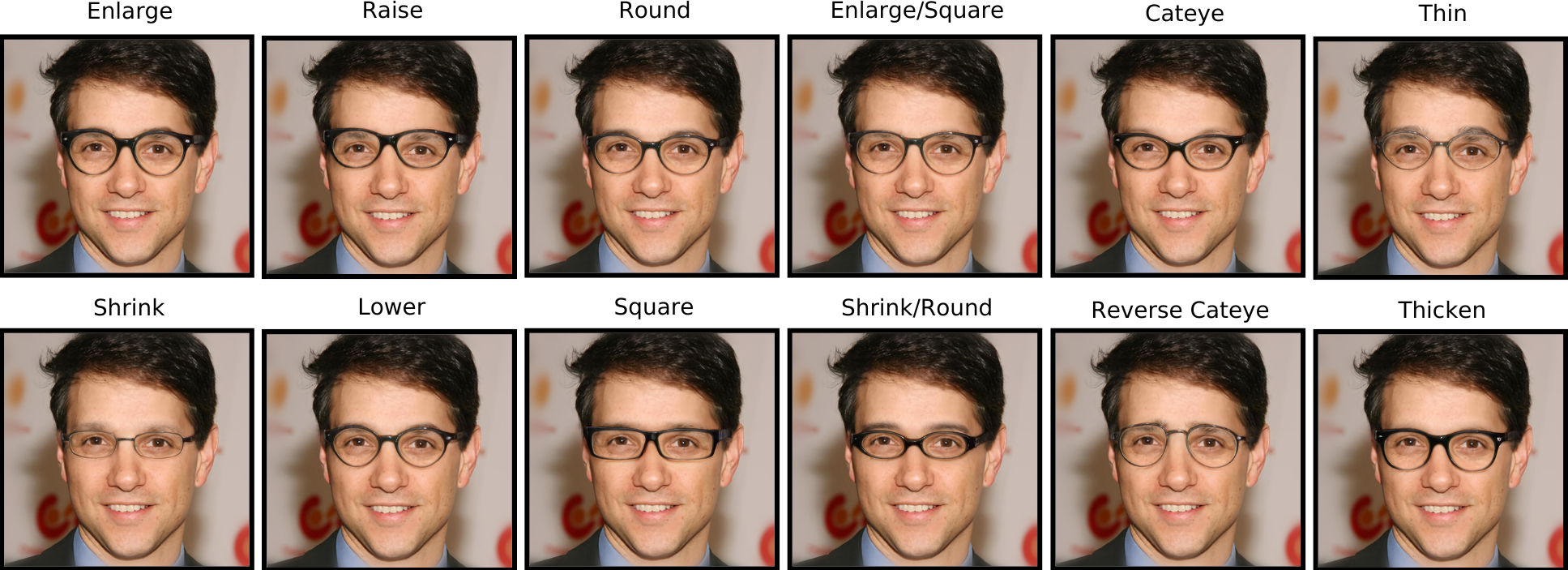}\vspace{-1mm}
        \figcaption{\small This paper introduces GlassesGAN, an innovative approach to image editing capable of generating continuously tunable, multi-attribute, and photo-realistic editing of eyeglasses by leveraging a novel method for modeling sub-spaces in the StyleGAN2 latent space. The presented ($1024\times1024$) examples show editing results for twelve different tuning attributes. Best viewed zoomed-in.\label{fig:cherry}}
    \end{minipage}
    \vspace{4mm} %
\end{minipage}

\vspace{-4mm}\begin{abstract}\vspace{-3mm}
We present GlassesGAN, a novel image editing framework for custom design of glasses, that sets a new standard in terms of image quality, edit realism, and continuous multi-style edit capability. To facilitate the editing process with GlassesGAN, we propose a Targeted Subspace Modelling (TSM) procedure that, based on a novel mechanism for (synthetic) appearance discovery in the latent space of a pre-trained GAN generator, constructs an eyeglasses-specific (latent) subspace that the editing framework can utilize. Additionally, we also introduce an appearance-constrained subspace initialization (SI) technique that centers the latent representation of the given input image in the well-defined part of the constructed subspace to improve the reliability of the learned edits. We test GlassesGAN on two (diverse) high-resolution datasets (CelebA-HQ and SiblingsDB-HQf) and compare it to three state-of-the-art competitors, i.e., InterfaceGAN, GANSpace, and MaskGAN. The reported results show that GlassesGAN convincingly outperforms all competing techniques, while offering additional functionality (e.g.,  fine-grained multi-style editing) not available with any of the competitors.  The source code will be made freely available. %
\end{abstract}

\section{Introduction}%
\label{sec:Introduction}

Consumers are increasingly choosing the convenience of online shopping over traditional brick and mortar stores~\cite{cheng2021fashion}. For the apparel industry–which traditionally relied on individuals being able to try on items to suit their taste and body shape before purchasing–the shift to digital commerce has created an unsustainable cycle of purchasing, shipping, and returns. Now, an estimated $85\%$ of manufactured fashion items end up in landfills each year, largely due to consumer returns and unsatisfied online customers ~\cite{remy2016style}. This problem highlights a greater need for tools to help consumers make better buying decisions before making %
purchases online. \vspace{-0.5mm}

In response to these challenges, the computer vision community has become increasingly interested in virtual-try-on (VTON) technologies~\cite{cheng2021fashion,gong2021aesthetics,han2018viton,fele2022c} that allow for the development of virtual fitting rooms. These tools give users the flexibility to “try-on” custom designs and personalize fashion items by creating (photo-realistic) images to help them in their decision making.\vspace{-0.5mm}

In recent years, considerable progress has been made in image-based virtual try-on techniques for clothing and apparel that do not require costly dedicated hardware and difficult-to-acquire 3D annotated data~\cite{han2018viton,fele2022c,dong2019towards,yang2020towards,jiang2022clothformer}. However, most deployed solutions for virtual eyewear try-on still largely rely on traditional computer graphics pipelines and 3D modeling\cite{marelli_faithful_2021,yuan_magic_2017,zhang_virtual_2017,azevedo_augmented_2016,niswar_virtual_2011}. Such solutions provide convincing results, but save for a few exceptions, e.g.,~\cite{huang_human-centric_2012}, are only able to handle predefined glasses and do not support custom designs and eyewear personalization. As for 2D data, relevant research for virtual eyewear try-on has mostly focused on editing technology (facilitated by Generative Adversarial Network - GANs~\cite{goodfellow_generative_2014}) capable of inserting glasses into an image~\cite{lee_maskgan_2020,press_emerging_2018,harkonen_ganspace_2020}. The images these methods generate are often impressive, but adding eyewear with finely tunable appearance control remains challenging. 

In this work, we address the need for a technology that allows users to experiment with customizable eyewear styles via photo-realistic images. Our solution, GlassesGAN, is an image editing framework that allows users to add and control the design of glasses to a diverse range of input images (at a high-resolution). Distinct from existing virtual try-on work in the vision literature, the goal of GlassesGAN is not to try-on existing glasses, rather to allow users to explore custom eyewear designs. This functionality can be immediately used by designers for exploring new ideas, or be paired with search functionality to help consumers find their ideal glasses. 

GlassesGAN is a GAN inversion method~\cite{xia2022gan}, that uses a novel \textit{Targeted Subspace Modeling (TSM)} technique to identify relevant directions within the latent space of a pre-trained GAN model that can be utilized to manipulate the appearance of eyeglasses in the edited images. A key component of GlassesGAN is a new  \textit{Synthetic Appearance Discovery (SAD)} mechanism. SAD samples GAN latent space for eyeglasses, without requiring real-world facial images with eyewear. Additionally, we propose an \textit{appearance-constrained subspace initialization procedure} for the (inference-time) editing stage, which produces more consistent results across a diverse range of input images. 
Our comprehensive evaluation of GlassesGAN\textemdash{}that included two diverse test datasets and comparisons to state-of-the-art solutions from the literature\textemdash{}yielded highly encouraging results.

In summary, our main contributions in this paper are as follows: \vspace{-1mm}
\begin{itemize}[noitemsep]

    \item We present GlassesGAN, an image editing framework for custom design of eyeglasses in a virtual try-on setting that sets a new standard in terms of output image quality, edit realism, and continuous multi-style edit capability, as illustrated in Figure~\ref{fig:cherry}.
    \item We introduce a Synthetic Appearance Discovery (SAD) mechanism and a Targeted Subspace Modeling (TSM) procedure, capable of capturing eyeglasses-appearance variations in the latent space of GAN models using glasses-free facial images only. 
    \item{We introduce a novel initialization procedure for the editing process that improves the reliability of the facial manipulations across different input images.}
\end{itemize}

\section{Related work}
\label{sec:RelatedWork}

In this section, we briefly review existing work needed to provide context for GlassesGAN. The reader is referred to some of the excellent surveys on generative models~\cite{wang2021generative}, image editing~\cite{xia2022gan, tolosana2020deepfakes} and virtual try-on~\cite{jong2020virtual,gong2021aesthetics,cheng2021fashion} for a more comprehensive coverage of relevant areas. 

\textbf{Generative Adversarial Networks (GANs)} represent a class of generative models capable of synthesizing realistic, high-quality imagery \cite{goodfellow_generative_2014} and consist of generative and discriminative sub-networks learned with competing  objectives \cite{goodfellow_generative_2014}. Recent advances in GAN design and associated training procedures have led to considerable progress in various areas, including image-to-image translation~\cite{isola2017image,wang2018high,zhu2017unpaired,kwon2021diagonal,pizzati2021comogan}, image attribute manipulation~\cite{harkonen_ganspace_2020,lee_maskgan_2020,shen2021interfacegan,xu2022transeditor,wang2022high} as well as virtual try-on and fashion-related applications~\cite{fele2022c,han2018viton,issenhuth2020not,ge2021parser}. Modern GAN models, such as StyleGAN (v1--v3)~\cite{karras2019style,karras2020analyzing,karras2021alias}, have had particular success in generating realistic high-resolution (facial) images and facilitate corresponding editing solutions.

\begin{figure*}[t]
\begin{center}
\vspace{-1mm}
  \includegraphics[width=1\textwidth, trim = 0 1cm 2.2cm 0cm, clip]{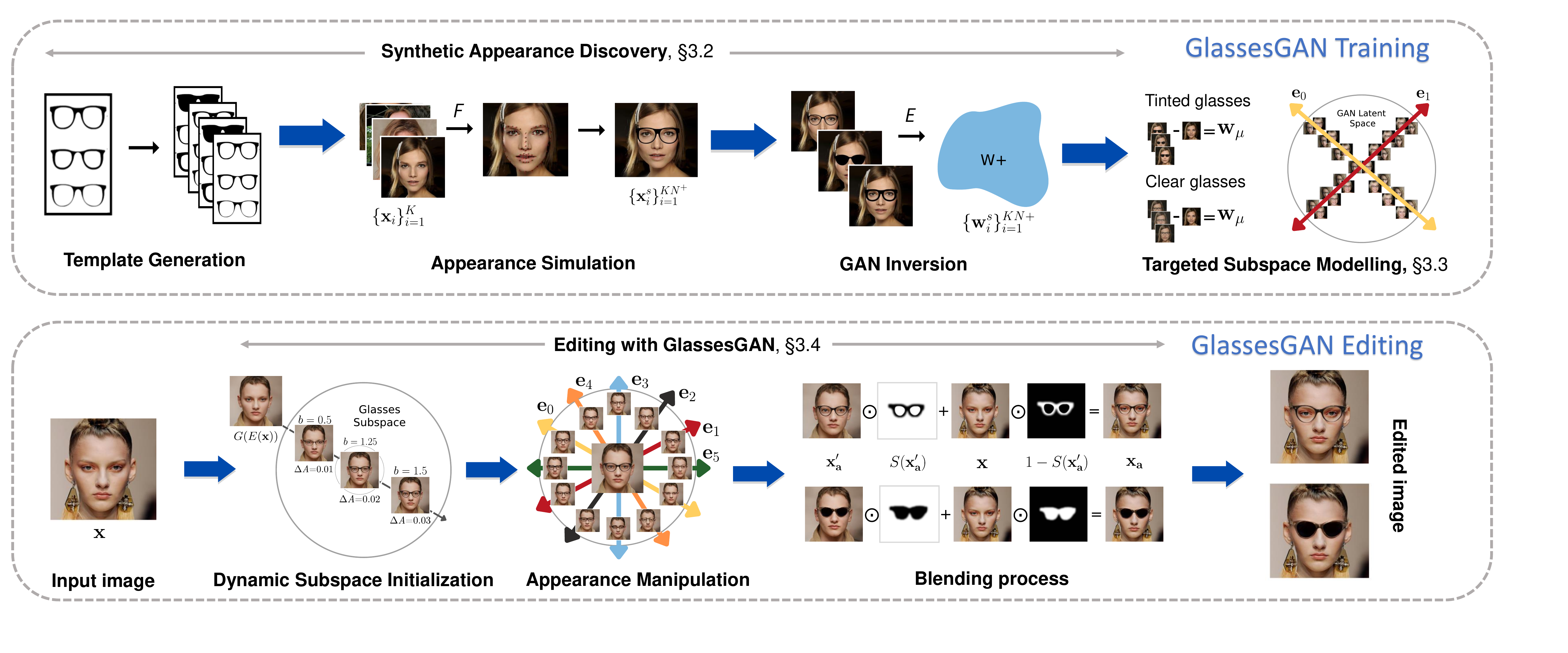}\vspace{-3mm}
    \caption{\textbf{Overview of the GlassesGAN framework.} GlassesGAN learns continuous multi-style edits through a novel GAN latent space sampling technique (synthetic appearance discovery) that first embeds augmented images into the latent space of a pretrained GAN generator and then captures the data distribution using the Karhunen-Lo\`eve Transform. During  editing, the framework dynamically initializes the latent vector in the center of the glasses subspace for greater edit consistency and then modifies different glasses attributes as desired.}\vspace{-7mm} %
  \label{fig:overview}
\end{center}
\end{figure*}

\textbf{Latent Space Image Editing} techniques
 alter attributes in the given input image by encoding the image in the GAN latent space, modifying the embedding, and then decoding the modified embedding~\cite{shen2021interfacegan,pernuvs2021high,khodadadeh2022latent,yang2021l2m,parmar2022spatially,liu2022towards}. While these types of methods can be very flexible, they typically suffer from a trade-off between editability, image consistency, distortion, and perceptual quality \cite{tov2021designing}. Additionally, the entanglement between different attributes in the generated images limits the locality of edits \cite{shen2021interfacegan,shen2020interpreting}. To mitigate such shortcomings, some researchers bypass the trade-off by blending the original image with the edited output image at strategic locations \cite{pernuvs2021high}, something we also follow with the proposed GlassesGAN framework in this work. 

\textbf{Glasses VTON.}
Recent works have had success creating VTON systems that rely upon detailed 3D modeling of the eyeglasses and/or the head \cite{marelli_faithful_2021,yuan_magic_2017,zhang_virtual_2017,azevedo_augmented_2016,niswar_virtual_2011,feng_virtual_2018, zhang_augmented_2018}. While some implementations have impressive edit realism, every additional eyeglass style (and person) requires a new 3D model. As a consequence, these techniques scale poorly to new eyeglasses, rarely contain the capability to make (continuous) edits to the eyeglasses, and sometimes require an initial 3D scan of the face/head to be applicable. 

To address such shortcomings, many recent methods try to avoid 3D data altogether and exploit advances in face image editing. These methods include latent space editing solutions, %
\cite{shen2020interpreting,harkonen_ganspace_2020,khodadadeh2022latent}, but also other editing strategies,
\cite{lee_maskgan_2020,he2019attgan}, capable of adding glasses to an input face image. While latent space editing techniques can be performed on 2D facial images and provide a realistic edit to the images, they have substantial problems with preserving identity throughout the edit, isolating the edit to the eyeglasses, and, prior to GlassesGAN, offered no multi-style personalization. 

\section{Methodology}\label{Sec: Methodology}

In this section, we present the main contribution of this work: GlassesGAN, a novel image editing framework %
that allows \textcolor{black}{for the personalization of eyeglasses in a virtual setting, i.e., with visual feedback to the user.} 

\subsection{Overview of GlassesGAN}

\textbf{Problem formulation.} Given an input face image $\mathbf{x}\in\mathbb{R}^{m \times n \times 3}$ and some desired semantics $a$ (i.e., appearance of glasses), the goal of GlassesGAN is to construct a mapping $\psi_a: \mathbf{x} \mapsto \mathbf{x}_a \in \mathbb{R}^{m \times n\times 3}$, such that the edited output image $\mathbf{x}_a$ incorporates the semantics $a$ in a realistic and visually convincing manner, while preserving the original image content as much as possible, e.g., facial appearance, background, and identity. A few illustrative examples of such edited images $\mathbf{x}_a$ are presented in Figure~\ref{fig:cherry}.  

Many state-of-the-art image-editing techniques implement the mapping $\psi_a$ through so-called GAN inversion approaches~\cite{xia2022gan,shen2021interfacegan}, where the input image $\mathbf{x}$ is first embedded into the latent space of a pretrained GAN generator $G$, thus, resulting in a latent representation $w$. This latent code is then modified, i.e., $\psi_a^{latent}:w\mapsto w_a$, such that the generated image $\mathbf{x}_a=G(w_a)$ adheres as closely as possible to the facial editing constraints. GlassesGAN follows this general latent-space editing framework, but in contrast to prior work:  $(i)$ does not require a dataset with the attribute $a$ present to define $\psi_a^{latent}$, and $(ii)$ learns latent space manipulations that enable \textit{continuous multi-style} changes to $a$. %

\textbf{GlassesGAN design.} A high-level overview of GlassesGAN in presented in Fig.~\ref{fig:overview}. Central to the editing ability of the framework are two novel components, i.e., $(i)$ a mechanism for \textit{Synthetic Appearance Discovery} (SAD) that allows us to sample target appearances of faces with various styles of glasses ($\mathbf{x}_a$) and their corresponding GAN latent codes \textbf{without actual real-world data}  (\secref{sec:SAD}), and $(ii)$ a \textit{Targeted Subspace Modeling} (TSM) approach (\secref{sec:TSM}) that based on the sampled representations, determines the latent editing directions using the Karhunen-Lo\`eve Transform. %

The identified latent directions correspond to different types of eyeglasses edits and can be applied to an input image's latent code as desired. To avoid problems with the latent space manipulations, we also propose a novel dynamic \textit{Subspace Initialization} (SI) procedure (\secref{sec:Inf}) that ensures that the generated edits are semantically meaningful. To produce the final output $\mathbf{x}_a$, we finally use a blending operation with the original image $\mathbf{x}$, which helps to preserve identity and to improve the locality of the edits.

\begin{figure}[t]
\begin{center}
  \includegraphics[width=0.99\columnwidth]{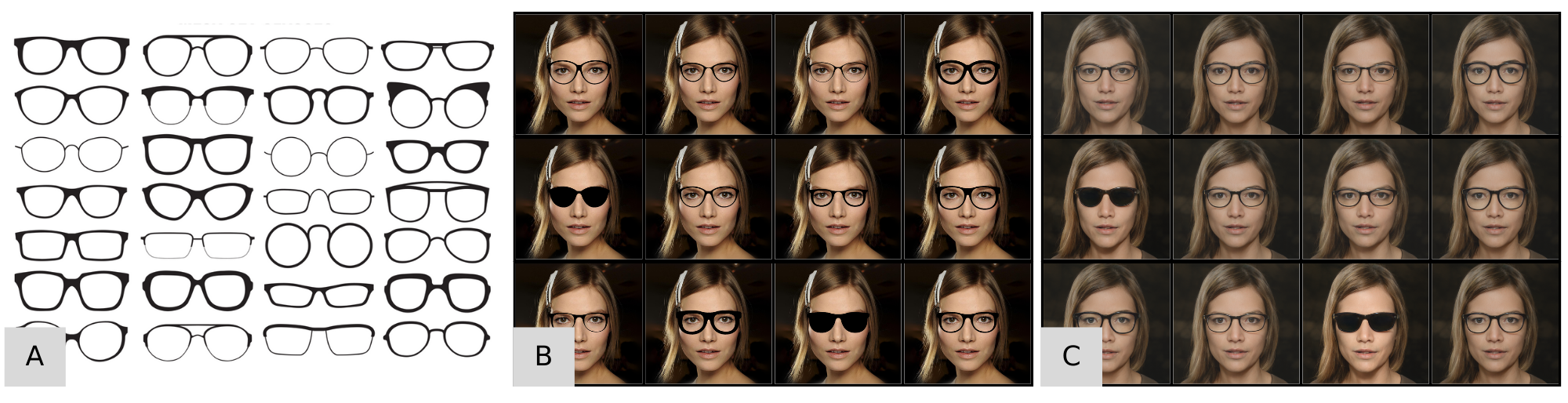}\vspace{-2mm}
    \caption{\textbf{Illustration of the SAD steps.} From left to right: (a) the initial (binary) glasses templates, (b) faces with superimposed templates, (c) re-renderings after latent space embedding. %
}\vspace{-5mm}
  \label{fig:SADprocedure}
\end{center}
\end{figure}

\subsection{Synthetic Appearance Discovery}\label{sec:SAD}

The majority of existing latent-space editing techniques require (paired or unpaired) data with and without the desired semantics $a$ to be able to learn the mapping $\psi_a$, e.g., \cite{shen2021interfacegan,yang2021l2m}. Since our goal is to provide fine-grained control over the appearance of glasses and suitable datasets for this purpose are not publicly available, we propose a Synthetic Appearance Discovery (SAD) mechanism  to mitigate this problem. Details on the mechanism are given below. %

\textbf{Step 1: Template generation.} We work under the assumption that only facial images $\mathbf{x}$ without glasses are available. To generate paired data with and without glasses, we simulate the presence of eyewear by superimposing hand-drawn binary masks $\mathbf{b}$ (\textit{glasses templates} hereafter) over the input images. We start this process with a collection of $N$ initial masks %
(see Figure~\ref{fig:SADprocedure}(a)), which we augment using morphological modifications, such as dilation and erosion, to expand the variability in the set of glasses templates. %

\textbf{Step 2: Appearance simulation.} Next, we add the augmented set of $N^+$ binary masks to each input image $\mathbf{x}$, resulting in facial images with an artificial cut-and-paste look $\mathbf{x}^s$, as illustrated in Figure~\ref{fig:SADprocedure}(b). The addition of the glasses is implemented based on a facial landmarking procedure $F$ that allows us to place the glasses templates on the faces in such a way that the temples of the head overlap with the outer points of the glasses frames %
(see also Figure~\ref{fig:overview}). Since $N^+$ glasses templates are available, this step results in a set of $N^+$ images $\{\mathbf{x}_i^{s}\}_{i=1}^{N^+}$ for each given input image. %

\textbf{Step 3: GAN inversion.} Finally, we embed the augmented images $\{\mathbf{x}_i^{s}\}_{i=1}^{N^+}$ in the latent space of the generator $G$  to obtain the corresponding latent codes $\{\mathbf{w}_i^{s}\}_{i=1}^{N^+}$. A pretrained StyleGAN2 model is used as the generator for GlassesGAN with the extended ($512\times18$ dimensional) $W^+$ latent space~\cite{karras2020analyzing}. %
We use an encoder-based approach for the GAN inversion, where the latent codes are computed as $\mathbf{w} = E(\mathbf{x})$ and $E$ represents the encoding operation. This last encoding step relies on the properties of pre-trained generator models, which are known to interpret image artifacts and binary occlusions in a semantically meaningful manner. As a result, the computed binary codes, simulate glasses with realistic appearance, and even add shadowing and specular reflections when re-rendered through the generator, i.e., $\mathbf{x}_s' = G(\mathbf{w}^s)$, as shown in Figure~\ref{fig:SADprocedure}(c).

If we assume a training set of $K$ glasses-free facial images, the SAD mechanism results in a dataset of $KN^+$ latent codes that capture the variability induced by the presence of eyeglasses and are used in the targeted subspace modeling (TSM) procedure, described in the next section.

\subsection{Targeted Subspace Modeling}\label{sec:TSM}

To facilitate continuous multi-style editing in the latent space, we introduce a Targeted Subspace Modeling (TSM) procedure, capable of identifying relevant latent space directions that, when traversed, result in visually meaningful modifications in the appearance of eyeglasses. Assume that: $(i)$ a training set of $K$ facial images without glasses is available, $(ii)$ that $N^+$ latent codes $\{\mathbf{w}_i^{s}\}_{i=1}^{N^+}$ have been computed with the SAD for each training image, and $(iii)$ that the center $\mathbf{w}_{\mu}^s=(1/N^+)\sum_{i=1}^{N^+}\mathbf{w}_i^s$ of these latent codes has been determined. TSM then first computes a differential latent code for each of the $K$ images, i.e.:
\begin{equation}
    \mathbf{\Delta W} = [vec(\mathbf{w}_1^s-\mathbf{w}_{\mu}^s), \ldots,vec(\mathbf{w}_{N^+}^s-\mathbf{w}_{\mu}^s)], %
\end{equation}
where $d$ is the dimensionality of the $W^+$ latent space (i.e., $d = 512\cdot18$) and $vec(\cdot)$ denotes a vectorization operator, and then aggregates the differentials over the training data:
\begin{equation}
    \mathbf{W} = [\mathbf{\Delta W}_1, \mathbf{\Delta W}_2, \ldots, \mathbf{\Delta W}_K]\in{\mathbb{R}^{d\times KN^+}}.
\end{equation}
The latent code differences in $\mathbf{W}$ capture the appearance variations of glasses, introduced to the training images by the SAD mechanism, and span a \textcolor{black}{\textit{glasses subspace}} 
within the latent space of the generator. As we show in the experimental section,  the differential formulation introduced above also allows us to model variations of different types of glasses (e.g., with clear and tinted lenses) using a single latent subspace. This subspace is identified by solving the eigenproblem given by  the Karhunen-Lo\`eve Transform: %
\begin{equation}
    \mathbf{\Sigma}\mathbf{e}_i = \lambda_i\mathbf{e}_i, \ \ i=1,2,\ldots, d',
    \label{eq:subspace_axes}
\end{equation}
where $\mathbf{\Sigma}=\mathbf{W}\mathbf{W}^T$ is an image-conditioned intra-class scatter matrix, and $d'\leq d$. The leading eigen-vectors corresponding to non-zero eigenvalues, i.e., $\mathbf{E}=[\mathbf{e}_1, \mathbf{e}_2, \ldots, \mathbf{e}_{d'}]\in\mathbb{R}^{d\times d'}$, define the (orthonormal) principal axes of the glasses subspace in $W^+$ and represent the basis for the image editing procedure of GlassesGAN. %

\textcolor{black}{As part of TSM, we also compute a difference vector $\mathbf{w}_\mu$ between the latent code $\mathbf{w}$ of each glasses-free training image and the centroid (i.e, mean vector) of the latent codes corresponding to the $KN^+$ glasses-augmented samples $\{\mathbf{w}_{i}^{s}\}_{i=1}^{KN^+}$. This difference vector is required for the initialization of the editing procedure.}

\subsection{Editing with GlassesGAN}\label{sec:Inf}

The editing procedure implemented for GlassesGAN consists of three main parts, as detailed below. %

\textbf{Part 1: Latent Code Editing.} 
Given a glasses-free input image 
$\mathbf{x}$ and its corresponding latent code $\mathbf{w}=E(\mathbf{x})$, computed with a pre-trained encoder $E$, we alter the initial latent code $\mathbf{w}$ by traversing the principal subspace axes in $\mathbf{E}$ using the following expression:
\begin{equation}
    \mathbf{w}_a' = vec(\mathbf{w}) + b\cdot vec(\mathbf{w}_\mu) + m\cdot\mathbf{e}_i,
    \label{eq:editing}
\end{equation}
where $i\in\{1,2,\ldots, d'\}$, $m\in[-\infty,\infty]$ is a real-valued scalar that controls the strength of the edits (\textit{editing magnitude} hereafter), $b$ is a weighting parameter that is set dynamically as part of the initialization procedure (described in Part 2), and the final $512\times 18$ latent representation $\mathbf{w}_a$ of the initial edited output image $\mathbf{x}_a'=G(\mathbf{w}_a)$ is computed as $\mathbf{w}_a = vec^{-1}(\mathbf{w}_a')$. Each principal axis $\{\mathbf{e}_i\}_{i=1}^{d'}$ controls a specific attribute (or style) of the glasses, while tuning the magnitude $m$ allows for continuous appearance changes w.r.t said attribute. %
The addition of the average difference code $\mathbf{w}_\mu$ serves as an initialization step that moves $\mathbf{w}$ into the well-defined part of the computed subspace, as shown in Figure~\ref{fig:overview}. 
If $\mathbf{w}_\mu$ is computed based only on latent codes corresponding to specific styles of glasses  (e.g., clear or tinted), then this code can also be used to define the initial appearance of the eyewear added to the image.

\textbf{Part 2: Dynamic Subspace Initialization.} 
It is important to note that the magnitude of the weighting parameter $b$ in Eq.~\eqref{eq:editing}, has significant downstream effects on later style edits, with improper values leading to eyeglasses that are poorly rendered or even non-existent. Similarly to prior work~\cite{pernuvs2021high,shen2021interfacegan}) we observed that the use of a fixed value of $b$ produces highly inconsistent edits across different samples. We hypothesize that this is because some samples are farther from the relevant part of the latent space than others. If the value of $b$ is too small for a particular sample, the embedding never enters the glasses subspace and, as a result, the glasses never (properly) appear. %
To address this issue, we propose a \textit{Subspace Initialization} (SI) procedure that dynamically adjusts the value of $b$ on a per-sample basis and ensures consistent editing results when using fixed, predefined style editing magnitudes $m$. Central to the initialization operation is the realization that the modified latent code $w_{a}$ is near the center of the glasses subspace when the frames of the glasses in the corresponds image $G(w_{a})$ cover a certain fraction $\Delta A$ of the overall image area. %
The initialization process therefore sets $m$ to zero, iteratively samples a range of values of $b$ from $0.5$ to $1.5$, generates an output image, subjects it to a face parser $S$ capable of segmenting the face from the glasses, and finally selects the optimal value of $b$, such that the frames in $G(w_{a})$ cover a relative  area as close to $\Delta A$ as possible, as shown in Figure~\ref{fig:overview}. %

\textbf{Part 3: Blending.} As illustrated in Figure~\ref{fig:overview}, in the last step, we finally blend the glasses region of the edited image $\mathbf{x}'_{a}$ with the original image $\mathbf{x}$ to improve the preservation of identity and, thus, compute the final output $\mathbf{x}_a$. The blending mask comes from the face parser $S$ applied to $x'_{a}$. The edges of the mask are tapered using Gaussian blur to smooth the boundary between the original and edited images. In the case of clear glasses two separate Gaussian blur operations for the interior and exterior of the glasses frames are used to better preserve the eyes of the original image.

\section{Experiments And Results} 
\label{sec:ExperimentalSetup}

In this section, we now present the experiments conducted to highlight the characteristics of GlassesGAN. %

\subsection{Datasets and Experimental Splits}

Three face datasets with diverse characteristics are used in the experiments with GlassesGAN, as summarized in Table~\ref{tab:datasets}, i.e.: FFHQ~\cite{karras2020analyzing}, CelebA-HQ~\cite{karras2018progressive}, and SiblingsDB-HQf~\cite{Vieira2014}. The datasets represent standard datasets used when evaluating image editing techniques and were, therefore, also selected for the experiments in this work~\cite{karras2020analyzing,pernuvs2021high}: %
\begin{itemize}[noitemsep]
    \item \textbf{FFHQ}~\cite{karras2020analyzing} consists of $70,000$ facial images of $1024\times1024$ pixels in size and was acquired from Flickr. Due to the unconstrained nature of the collection procedure, the datasets exhibits variability across various factors. FFHQ is used to train the generator $G$ and image encoder $E$ in our experiments.
    \item \textbf{CelebA-HQ}~\cite{karras2018progressive} contains high-quality facial images at a resolution of $1024\times1024$ pixels with considerable appearance variability. $1000$ sampled images are used for training (with SAD and TSM) in our experiments, and a non-overlapping subset of $1000$ diverse test images is used for the quantitative  evaluation. %
    \item \textbf{SiblingsDB-HQf}~\cite{Vieira2014} contains $184$ frontal facial images of $92$ sibling pairs captured at a resolution of $4256\times2832$. The dataset was acquired in front of a homogenous background and under diffuse illumination. This dataset is  used exclusively for testing to demonstrate the generalization capabilities of GlassesGAN across datasets. After removing duplicates and excluding problematic samples, $163$ image are left for the quantitative part of the evaluation. \vspace{-1mm}
\end{itemize}
We note that the training and test data is kept disjoint in all experiments, both in terms of images and subjects identities.

\begin{table}[!t!]
\centering
\caption{\textbf{Summary of the experimental datasets and data splits.\vspace{-2mm}}}
    \label{tab:datasets}
\renewcommand{\arraystretch}{1.1}
\resizebox{\columnwidth}{!}{%
\begin{tabular}{lccccc}
\toprule
\textbf{Dataset} & 
\textbf{Resolution} & \textbf{Purpose$^\diamond$}  & 
\textbf{\#Train. Img.}$^\dagger$ &
\textbf{\#Test Img.} & \textbf{Variability$^\ddagger$}\\ \midrule
FFHQ~\cite{karras2020analyzing} & 
$1024\times 1024$ & 
TR ($G$,$E$)  & 
$70,000$ & 
n/a & 
A, ET, G, B\\
CelebA–HQ~\cite{karras2018progressive}  &
$1024\times 1024$ & 
TR ($S$), Q, TS & 
$1000$ & 
$1000$ & 
A, ET, B, G, AC\\
SiblingsDB-HQf~\cite{Vieira2014} & 
$4256\times 2832$ & 
Q, TS & 
n/a & 
$163$ & 
A, G\\
\bottomrule
\multicolumn{6}{l}{$^\diamond$ TR -- training, Q -- qualitative evaluation, TS -- quantitative evaluation (testing), n/a -- not applicable.}\\
\multicolumn{6}{l}{$^\dagger$ The number of training images reported includes both training and validation data.}\\
\multicolumn{6}{l}{$^\ddagger$ A -- age, ET -- ethnicity, G -- gender, B -- background, AC -- accessories.}\\
\end{tabular}
}\vspace{-3mm}
\end{table}

\subsection{Implementation Details and Runtime}

For the implementation of GlassesGAN, we use StyleGAN2 at resolution $1024\times1024$ trained on images from FFHQ~\cite{karras2020analyzing} as the generator $G$ of our framework and the e4e~\cite{tov2021designing} encoder $E$ again trained on FFHQ for inverting images into StyleGAN's latent space. For face detection and identifying facial landmarks we utilize the $68$-point landmark model provided in the dlib package. %
We adopt the DatasetGAN~\cite{zhang2021datasetgan} framework with $7$ manual annotated data samples to generate synthetic training data for the face parser and then learn a DeeplabV2 model to serve as the parser $S$ in our experiments \cite{chen2017deeplab}. We construct a $d'=6$ dimensional subspace from the CelebA-HQ training images, and use $N=28$ glasses templates for TSM. The image area threshold $\Delta A$ is set to $0.02$ based on preliminary experiments. %
With the current implementation using an RTX $3090$ GPU, adding glasses and applying an edit to an image requires $4.8$s on average (estimated over $100$ test images). The addition of the subspace initialization to the pipeline costs an additional $12.7$s. %
However, an efficient parallel implementation of GlassesGAN is expected to allow for real-time editing capability. Additional implementation details can be found in the publicly released source code.

\begin{figure}[t]
\begin{flushleft}
\end{flushleft}
\vspace{-7mm}
\begin{center}
  \includegraphics[width=1.0\linewidth, trim = 0 0 0 5mm, clip]{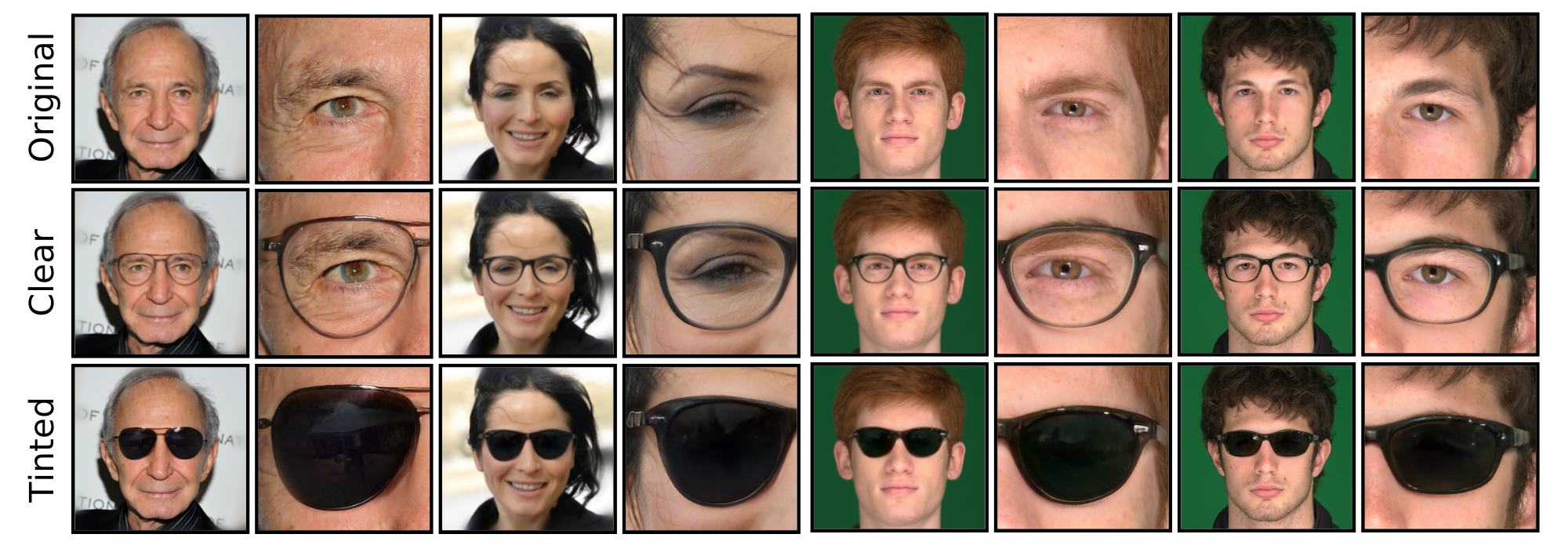}\vspace{-2mm}
    \caption{\textbf{Addition of initial glasses of a certain style.} GlassesGAN is able to render and edit glasses in different styles. The bottom two rows shows examples of the initialization with (average/initial) clear (middle) and tinted (bottom) glasses added to the (CelebA-HQ and SiblingsDB-HQf) input images on the top.}\vspace{-5mm}
  \label{fig:avg_glasses}
\end{center}
\end{figure}

\subsection{Qualitative Results}

To demonstrate the capabilities of GlassesGAN, we first present a series of visual results that illustrate: $(i)$ the addition of two different types of initial eyeglasses to a face image, $(ii)$ the tuning of eyeglasses appearance with respect to different attributes, $(iii)$ sequentially chaining of eyeglass style edits, and $(iv)$ edits to eyeglass frame color. %

\textbf{Adding initial glasses.} The initialization procedure of GlassesGAN requires that a starting point is chosen in the glasses subspace via $\mathbf{w}_\mu$ in Eq.~\eqref{eq:editing}. This starting point defines the initial appearance and shape of the rendered glasses and can be varied to achieve different results, i.e., different initial styles of glasses. In Figure~\ref{fig:avg_glasses} we show a number of qualitative examples, where initial (average) clear and tinted glasses were added to the input images.  
As can be seen, GlassesGAN is able to add glasses to input images with diverse appearances (i.e., varying gender, age, background, color characteristics, etc.) and automatically consider facial alignment, shadowing, the boundary with the hair, and reflections in the frames and lenses. %
Additionally, we see that the blending procedure helps to maintain the fine image details while still preserving identity.

\begin{figure}[t]
\begin{center}\vspace{-2mm}
  \includegraphics[width=0.89\linewidth]{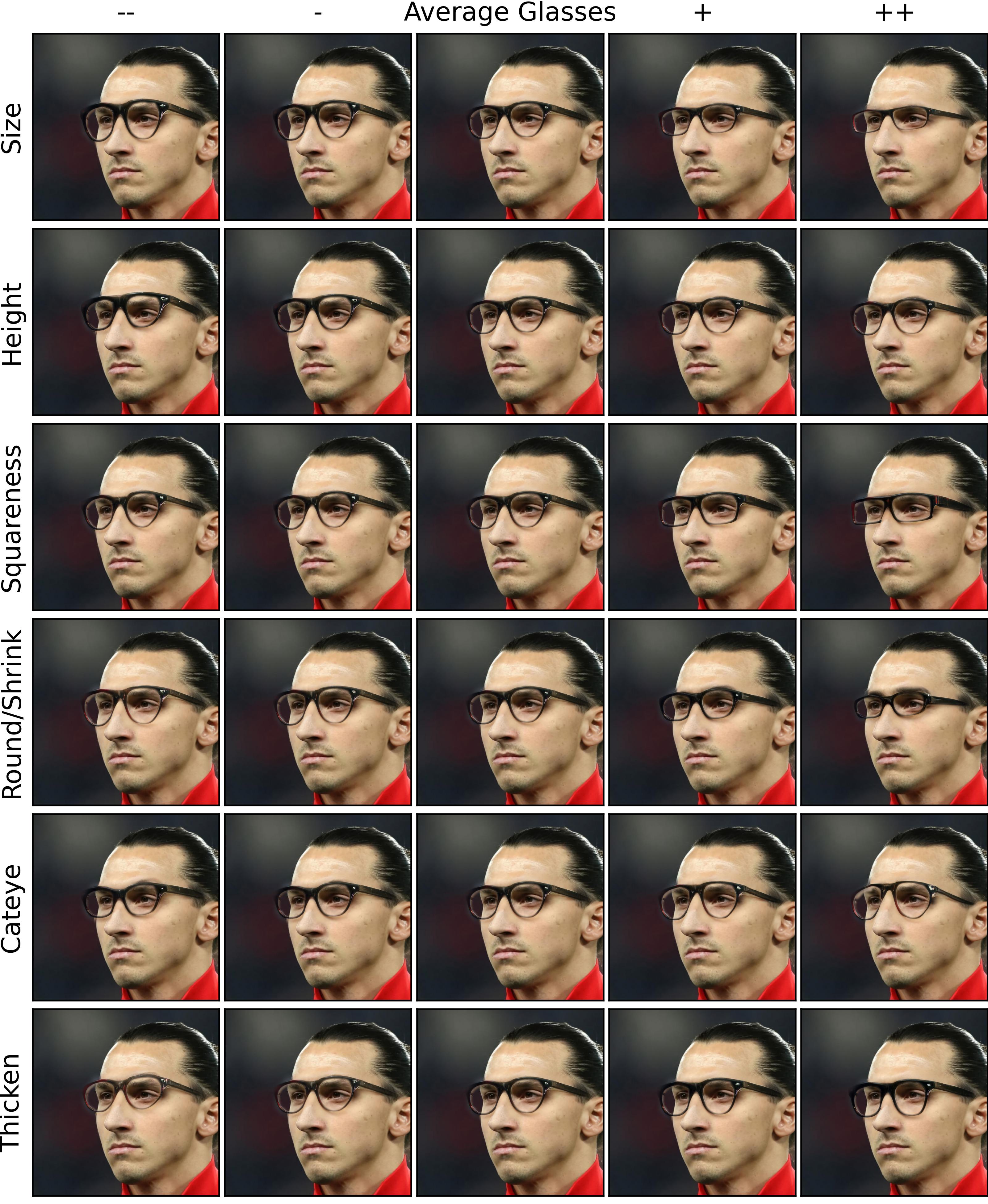}\vspace{-2mm}
    \caption{\textbf{Continuous multi-style edits.} Each row shows a separate edit on a challenging pose that starts from the (initialized) image in the middle and modifies one aspect of the glasses in a given direction. Results corresponding to the first six subspace axes (top to bottom) identified through the TSM procedure are presented.%
    }\vspace{-5mm}
  \label{fig:continuous_multifactor_edit_singlecol}
\end{center}
\end{figure}

\textbf{Editing different attributes.} Using the TSM procedure, GlassesGAN identifies a number of latent subspace directions that can be traversed to alter the appearance of the generated glasses. In Figure~\ref{fig:continuous_multifactor_edit_singlecol} we present a few visual examples where the initial glasses in the middle column are altered (continuously) in six different directions. Each of the rows corresponds to changes along one subspace axis from Eq.~\eqref{eq:subspace_axes}. Because the TSM procedure is unsupervised, we subjectively assign human-interpretable attributes to these directions, which impact the following aspects of the added glasses: %
size, height/position, squareness, roundness, cat-eye appearance, and thickness. Note that each of the edits is visually convincing and creates distinct appearances. %

\begin{figure}[t]
\begin{center}
  \includegraphics[width=0.95\linewidth]{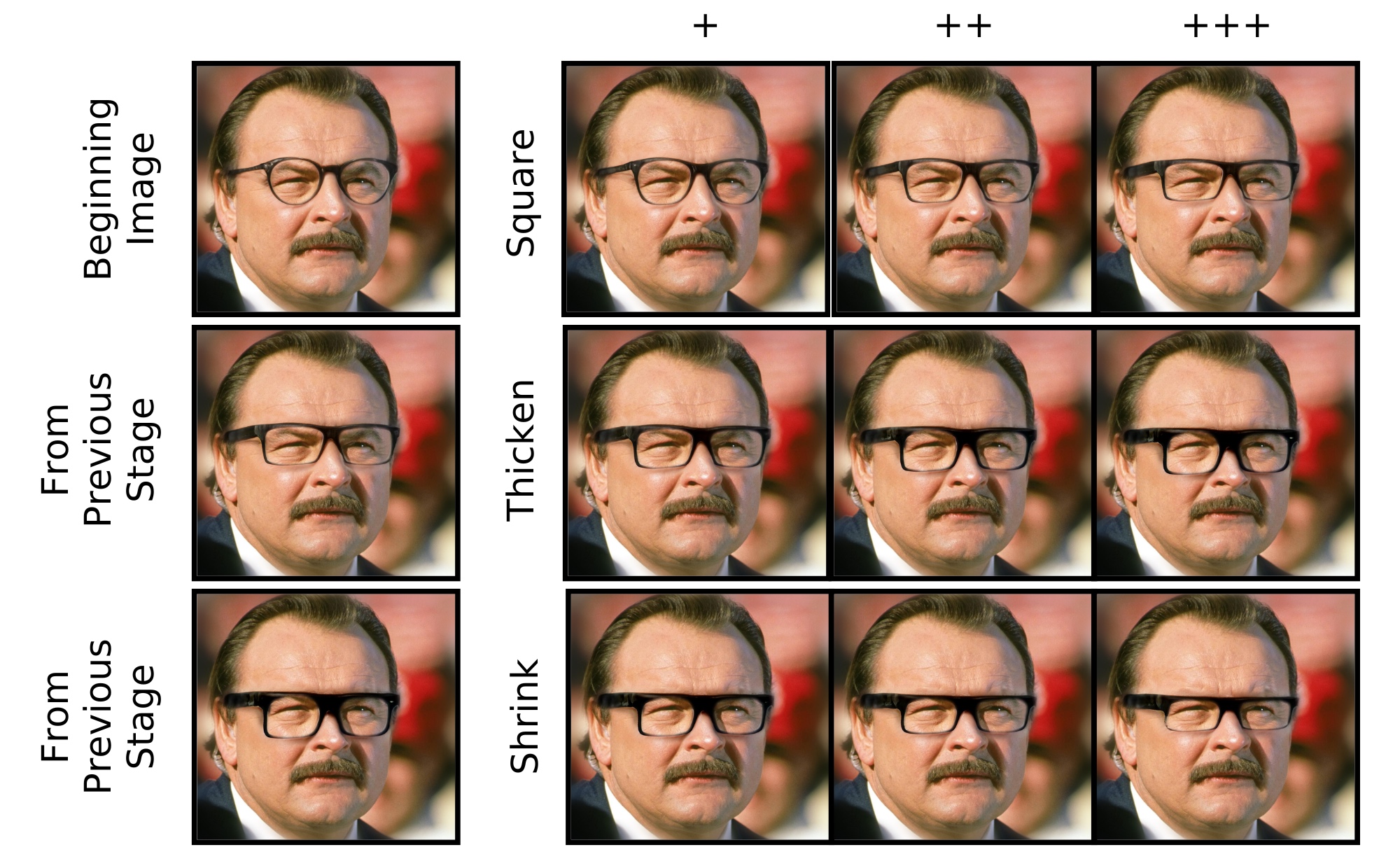}\vspace{-2mm}
    \caption{\textbf{Example of sequentially chained edits}. GlassesGAN allows to chain edits in different latent subspace directions without affecting the realism of the  results or introducing artifacts.} %
    \vspace{-5mm}
  \label{fig:multi_direction_edit}
\end{center}
\end{figure}

\textbf{Multiple chained edits.} Next, we show that the latent subspace directions exploited by GlassesGAN are disentangled enough (due to the orthogonality of the learned subspace) to allow multiple chained edits to an image. For example, in Figure~\ref{fig:multi_direction_edit}, we show that eyeglasses can be sequentially squared, thickened, and then shrunk. This chained editing procedure allows for the generation of unique appearances of glasses and fine-grained control over the editing procedure - a  characteristic unique to our framework. %

\begin{figure}[t]
\begin{center}
  \includegraphics[width=1.0\linewidth]{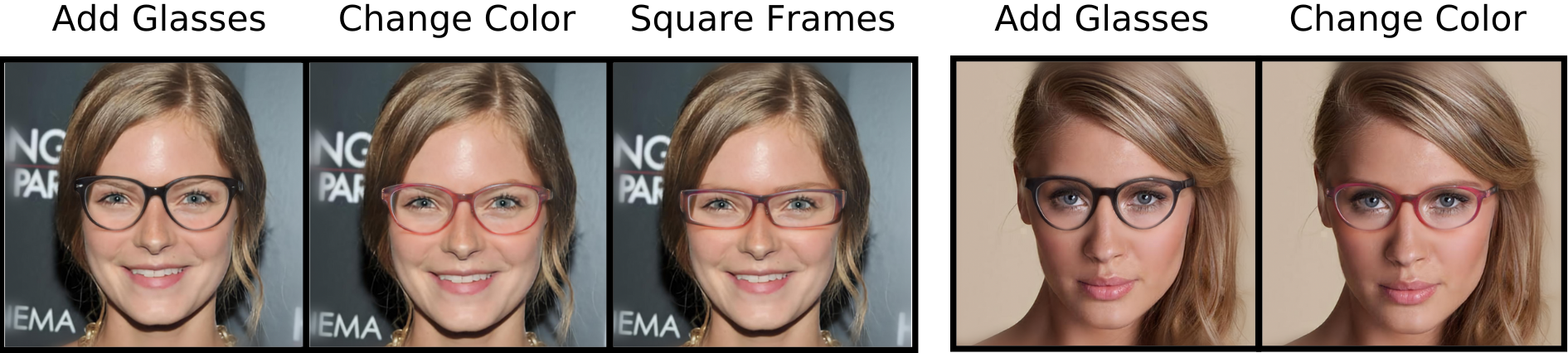}\vspace{-2mm}
    \caption{\textbf{Examples of color-related edits with GlassesGAN.} %
    The example of the left adds glasses to the face, changes the color, and then squares frames using the learned edit vectors. The right example first adds the glasses and then changes the color.} \vspace{-4mm}
  \label{fig:color_change}
\end{center}
\end{figure}

\textbf{Color change.} \textcolor{black}{In Figure \ref{fig:color_change} we demonstrate the ability of GlassesGAN to also capture attributes beyond the shape of the glasses. Specifically, by using colored augmentations for the glasses templates used in the SAD mechanism, we obtain frame-color control that is (reasonably well) disentangled from our suite of frame shape edits. This speaks of the flexibility of the framework and points to the potential for supporting further editing attributes if required.}

\begin{figure}[t]
\begin{center}
  \includegraphics[width=0.9\linewidth]{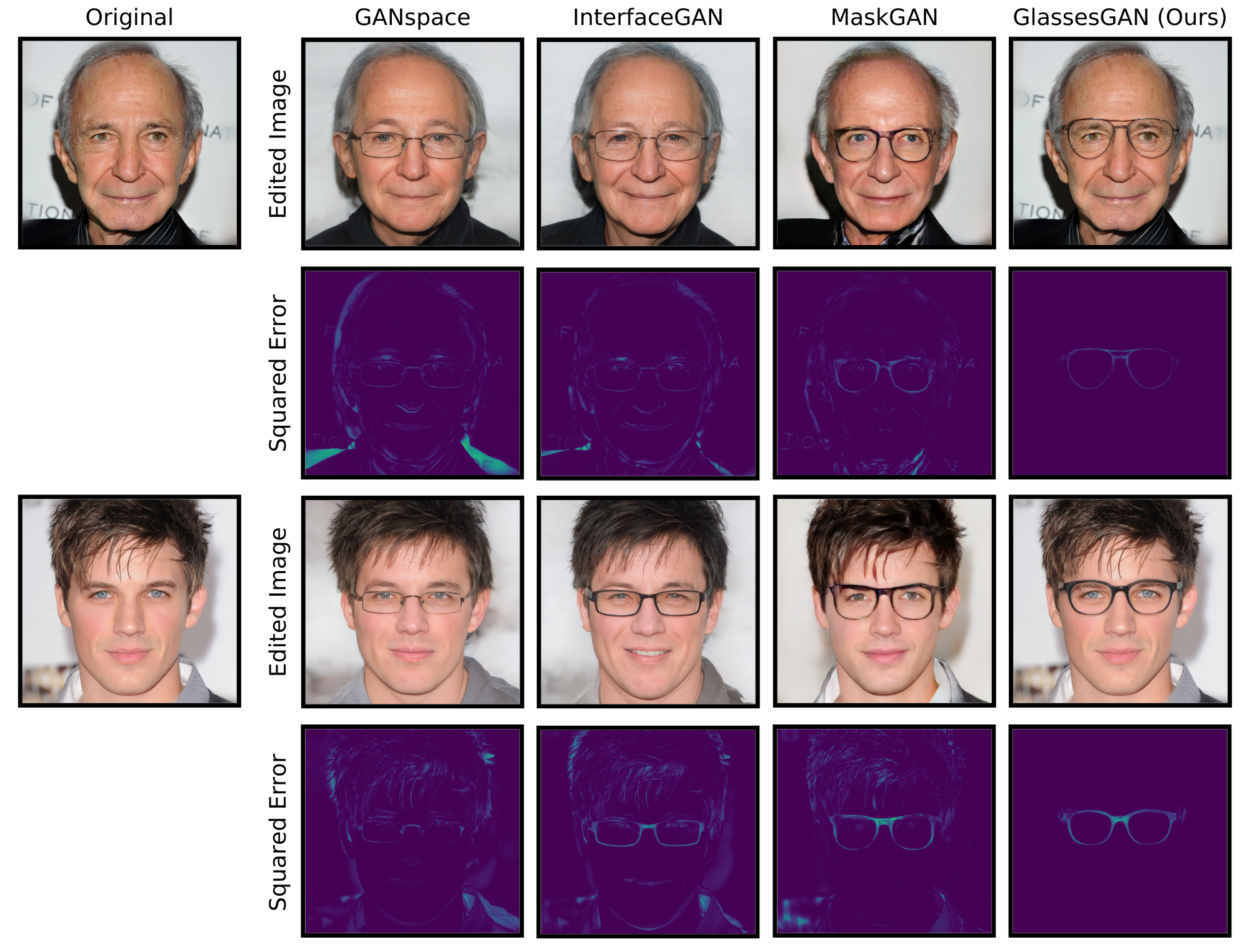}\vspace{-2mm}
    \caption{\textbf{Comparison to the state-of-the-art.} The examples show that GlassesGAN generates convincing results with minimal (or no) changes in identity. Squared pixel differences between the originals and edits are shown to highlight modified image areas.}\vspace{-3mm}
  \label{fig:squared_error_comparison}
\end{center}
\end{figure}

\subsection{Comparison to the State-Of-The-Art}

We compare GlassesGAN to three (related) state-of-the-art  image-editing techniques utilizing GANs, i.e.: InterFaceGAN~\cite{shen2021interfacegan}, MaskGAN~\cite{lee2020maskgan} and GANSpace~\cite{harkonen2020ganspace}. We note that the overall objective of GlassesGAN (i.e., custom design of glasses with visual feedback) is distinct, so \textbf{no direct competitors are available} in the literature. We, therefore, select the listed methods as our baselines, as they are able to add glasses (on/off) to facial images and (in some cases) ensure limited amounts of appearance control. 

\textbf{Visual comparison.} In Figure~\ref{fig:squared_error_comparison} we present results with a couple of \textcolor{black}{left-out} test images from the CelebA-HQ dataset 
To ensure a fair comparison, we use publicly released code for the baselines and set the hyperparameters in a way that ensures optimal visual results. Additionally, we select a discrete setting (i.e., appearance of glasses) for GlassesGAN that results in the addition of glasses similar to those produced by the competing methods. As can be seen from the results, all methods generate realistic eyeglasses, but except for GlassesGAN also introduce significant identity changes. While this is a common issue with latent-space based techniques, our framework avoids such problem through the use of blending, which leads to excellent edit
locality compared to the baselines. In contrast, the baseline methods introduce undesirable global changes
to the facial appearance, as also highlighted by the squared pixel differences in Figure~\ref{fig:squared_error_comparison}. 

\begin{table}[t]
\centering
\caption{\textbf{Comparison to the state-of-the-art.} GlassesGAN outperforms all baselines on both test datasets across nearly all performance indicators by a wide margin. The arrow ($\downarrow\uparrow$) indicates if lower or higher scores imply better performance.}\vspace{-2mm}
\label{tab:comparitivemetrics}
\resizebox{\linewidth}{!}{
\begin{tabular}{lrrr}  %
\toprule
\multirow{2}{*}{\textbf{Method}}
& \multicolumn{3}{c}{\textbf{CelebA-HQ}} \\ \cline{2-4}

& MSE ($\downarrow$) 
& IDS ($\downarrow$) 
& FID ($\downarrow$) 
\\ \midrule

{InterfaceGAN~\cite{shen2021interfacegan}} 
& $0.0173 \pm 0.0058$ 
& $0.5789 \pm 0.1026$ 
& $58.15$  
\\

{MaskGAN$^\dagger$~\cite{lee2020maskgan}} 
& $0.0149 \pm 0.0064$ 
& $0.6568  \pm 0.0975$ 
& $53.11$  
\\

{GANSpace~\cite{harkonen2020ganspace}} 
& $0.0153 \pm 0.0064$ 
& $0.4842 \pm 0.1060$ 
& $40.45$
\\

{GlassesGAN (ours)} & 
$\boldsymbol{0.0029 \pm 0.0009}$ 
& $\boldsymbol{0.1707 \pm 0.0625}$ 
& $\boldsymbol{26.02}$
\\
\bottomrule

\multirow{2}{*}{\textbf{Method}}
& \multicolumn{3}{c}{\textbf{SiblingsDB-HQF}} \\ \cline{2-4}

& MSE ($\downarrow$) 
& IDS ($\downarrow$) 
& FID ($\downarrow$) 
\\ \midrule

{InterfaceGAN~\cite{shen2021interfacegan}} 
& $0.0099 \pm 0.0022$ 
& $0.5780  \pm 0.0790$ 
& $80.64$
\\

{GANSpace~\cite{harkonen2020ganspace}} 
& $0.0085 \pm 0.0030$ 
& $0.5047 \pm 0.0883$ 
& $60.79$ 
\\

{GlassesGAN (ours)}  
& $\boldsymbol{0.0029 \pm 0.0007}$ 
& $\boldsymbol{0.1589  \pm 0.0478}$ 
& $\boldsymbol{45.83}$ 
\\
\bottomrule
\multicolumn{4}{l}{$^\dagger$Requires a specific segmentation map not available for SiblingsDB-HQf.}
\end{tabular}
}%
\end{table}

\begin{figure}[t]
\begin{flushleft}
\end{flushleft}
\vspace{-6mm}
\begin{center}
  \includegraphics[width=0.56\linewidth, trim = 0 0mm 0 1mm, clip]{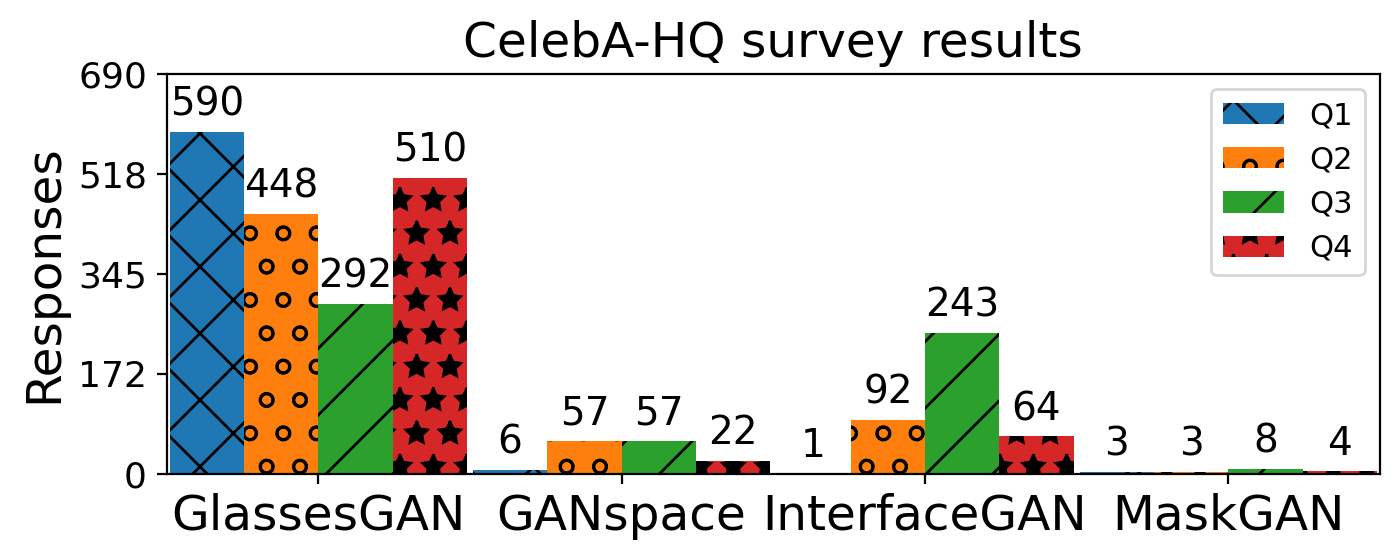}\hfill
  \includegraphics[width=0.44\linewidth, trim = 0 0mm 0 1mm, clip]{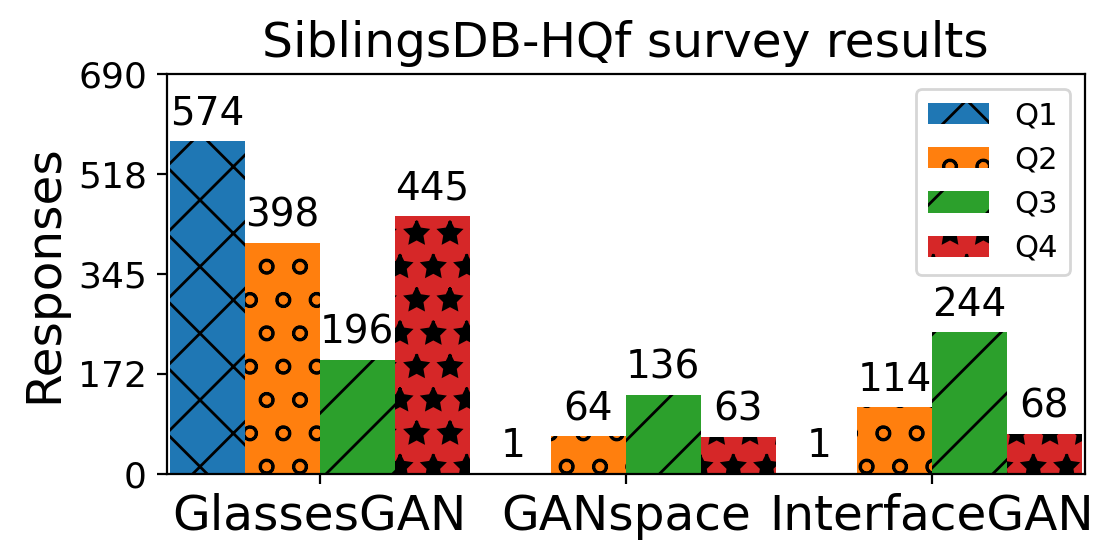}
  \vspace{-5mm}
    \caption{\textbf{User-study results.} Raters were asked to choose the method with the best identity preservation (Q1), eyeglasses quality (Q2), realism (Q3), and overall try-on result (Q4). Note that MaskGAN requires a specific segmentation map not available for SiblingsDB-HQf and is, therefore, not included in the right graph.}\vspace{-5mm}
  \label{fig:human_study}
\end{center}
\end{figure}

\textbf{Quantitative comparison.}
Next, we perform a quantitative comparison with the state-of-the-art on the designated (left-out) images from the CelebA-HQ and SiblingsDB-HQf datasets. We add glasses to the test images, using a similar procedure as for the visual comparison discussed above. Following established evaluation methodology~\cite{lee_maskgan_2020,pernuvs2021high}, we analyze the results in Table~\ref{tab:comparitivemetrics} from \textcolor{black}{four} different perspectives: $(i)$ through  \textit{Mean Square Error} (MSE) scores, computed between the original and edited images, to quantify unwanted pixel-level changes in the edited images, $(ii)$ through \textit{Identity Discrepancy Scores} (IDS), measured with Euclidean distances of the embeddings produced by a pre-trained ArcFace model~\cite{deng2019arcface} from the original and edited samples, to capture potential identity changes, introduced by the editing, $(iii)$ through Fréchet Inception Distances (FID) \cite{heusel2017gans} with the original input samples that reflect the realism and quality of the edited images, \textcolor{black}{and $(iv)$ through a user study with $4,704$ responses from $12$ human evaluators. For the study, evaluators were shown randomly selected test images and randomly ordered edits from each method and asked to choose the best identity preservation (Q1), quality (Q2), realism (Q3), and overall try-on result (Q4).}

From the results in Table~\ref{tab:comparitivemetrics} and Figure~\ref{fig:human_study}, we observe that GlassesGAN leads to significantly lower MSE scores on both datasets, suggesting that the GlassesGAN edits are closest to the originals among all tested methods. 
Our approach also has substantially less identity drift from the editing process, as shown by the IDS scores that are lower by a factor of $3$ compared to the closest competitor and the average user preference of $99\%$ on Q1. 
Additionally, the edited images generated by GlassesGAN result in the highest perceptual similarity to the original samples among all tested methods, as evidenced by the lowest observed FID scores (see Table~\ref{tab:ablations} for results without blending). Finally, the user survey results in Figure~\ref{fig:human_study} show a general user preference for GlassesGAN over the baseline methods. %

\begin{figure}
\vspace{-4mm}
\captionof{table}{\textbf{Ablation-study results.} The left table shows results with (w) and without (w/o) image blending, and the right with (w) and without (w/o) subspace tuning.\label{tab:ablations}}\vspace{1mm}
\resizebox{0.58\linewidth}{!}{
\begin{tabular}{lrrr}
\toprule
\textbf{GlassesGAN} 
& \multicolumn{3}{c}{\textbf{CelebA-HQ}} \\ \cline{2-4}

\textbf{Version} 
& MSE ($\downarrow$) 
& IDS ($\downarrow$)
& FID ($\downarrow$)\\ \midrule

{w/o Blending} 
& $0.0161$ 
& $0.6231$ 
& $80.72$ \\ 

{w Blending} & 
$\boldsymbol{0.0029}$ 
& $\boldsymbol{0.1707}$ 
& $\boldsymbol{26.02}$ 
\\ \bottomrule
\end{tabular}
}
\hfill
\resizebox{0.38\linewidth}{!}{
\begin{tabular}{lc}
\toprule
\textbf{GlassesGAN} 
& {\textbf{CelebA-HQ}} \\ \cline{2-2}
\textbf{Version} 
& ERS [in \%]\\ \midrule
{w/o Tuning} 
& $29.12\%$ \\ 
{w Tuning} & 
$\boldsymbol{6.22\%}$ 
\\ \bottomrule
\end{tabular}
}
\vspace{-2mm}
\end{figure}

\subsection{Ablation Studies}

We present ablation studies that investigate the impact of $(i)$ image blending and $(ii)$ subspace initialization.%

\begin{figure}[t]
\begin{center}
  \includegraphics[width=1.0\linewidth]{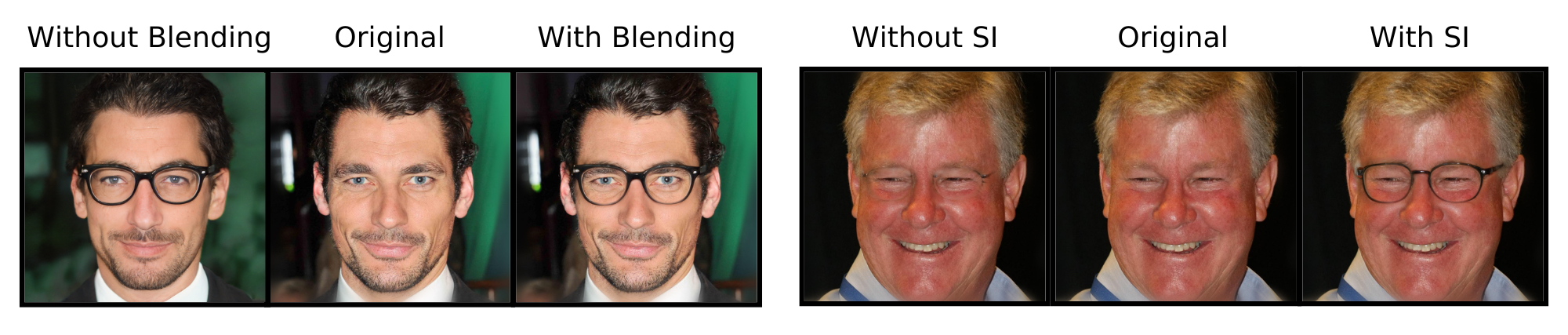}\vspace{-2mm}
    \caption{\textbf{Visual ablation-study results.} The left images show sample results with (w) and without (w/o) image blending, and the right w and w/o subspace initialization.}\vspace{-4mm}
  \label{fig:ablation}
\end{center}
\end{figure}

\textbf{Image blending.} The purpose of image blending is to improve the preservation of the subjects' identity and the locality of the edits. Focusing on image identity first, we compare the average IDS scores between the original and edited images with (w) and without (w/o) blending. As we show on the left part of Table \ref{tab:ablations}, blending substantially reduces the average identity discrepancy. Furthermore, it also reduces the MSE scores by more than $5\times$  and the FID scores by more than $3\times$. These results are further supported by the visual results on the left of Figure~\ref{fig:ablation}, where blending is again seen to have a beneficial effect on the editing output.%

\textbf{Subspace initialization.} The subspace initialization procedure is designed to improve the robustness of glasses manipulations by normalizing the latent-space edits dynamically on a per-sample basis to constrain the edited latent embeddings to the well-defined part of the learned subspace. To quantify the effectiveness of this solution, we develop a performance measure, we refer to as \textit{Edit Robustness Score} (ERS). ERS is defined as the probability that a latent-space edit fails because it is not conducted within the relevant edit space, which in turn leads to editing outputs without glasses. Failed edits with missing glasses are identified with a face parser ($S$) based on the area of the glasses frames, and the failure probability is estimated on the test images of CelebA-HQ. As can be seen from the right part of Table~\ref{tab:ablations}, the subspace initialization helps to reduce ERS scores by a factor close to $5\times$ and makes the editing process significantly more consistent. This can also be seen from the visual example on the right of Figure~\ref{fig:ablation}.

\begin{figure}[!t!]
\begin{center}
  \includegraphics[width=1\linewidth]{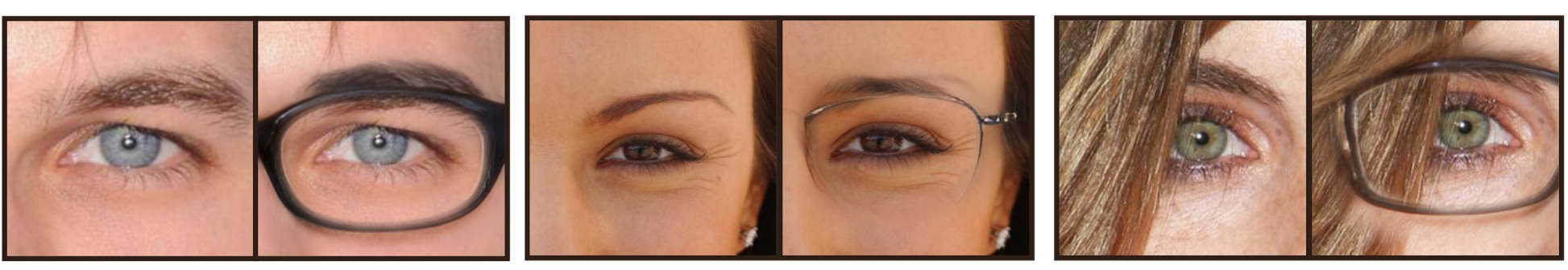}\vspace{-1mm}
    \caption{\textbf{Illustration of limitations.} Each presented pair shows the original image on the left and the edited one on the right. Parser errors, occlusions, and unusual image characteristics are the main causes of weaker results with a small fraction of the test images. %
    }\vspace{-5mm}
  \label{fig:limitations}
\end{center}
\end{figure}

\subsection{Limitations}

In Figure~\ref{fig:limitations}, we present some limitations of GlassesGAN. Because the framework relies on a face parser $S$, parser errors may affect the visual quality of the generated results. \textcolor{black}{In the left most example, we see that eyebrows are segmented as part of the frames, leading to changes in appearance. In the middle example, an incorrectly estimated frame area results in improper subspace tuning and poorly visible glasses. In the right, we see that rare occlusions by hair may result in glasses rendered in front of instead of behind the hair. While the visual quality of these examples is still reasonable, such errors are expected to benefit from future advancement in the auxiliary models, e.g., parser $S$.}

\section{Conclusion}%

In this paper, we presented GlassesGAN, a framework for facial image editing that allows the addition of different styles of glasses to input images and continuous editing of their appearance. Extensive experiments over diverse test datasets showed that GlassesGAN yields convincing edits across images with rich appearance variations, while comparing favorably to competing methods. Even though the framework was designed
\textcolor{black}{to allow custom creation of glasses,} facial editing technology in general may also have \textit{unintended negative social impact} as the modified images could be misused for public shaming, fraud, and manipulating public option. Proper safeguards, therefore, need to be taken when deploying such technology in practice. %

{\small
\bibliographystyle{ieee_fullname}
\bibliography{egbib}

\begin{thebibliography}{10}\itemsep=-1pt

\bibitem{azevedo_augmented_2016}
Pedro Azevedo, Thiago~Oliveira Dos~Santos, and Edilson De~Aguiar.
\newblock An augmented reality virtual glasses try-on system.
\newblock In {\em Symposium on Virtual and Augmented Reality ({SVR})}, pages
  1--9, 2016.

\bibitem{chen2017deeplab}
Liang-Chieh Chen, George Papandreou, Iasonas Kokkinos, Kevin Murphy, and Alan~L
  Yuille.
\newblock {Deeplab: Semantic image segmentation with deep convolutional nets,
  atrous convolution, and fully connected CRFs}.
\newblock {\em IEEE Transactions on Pattern Analysis and Machine Intelligence
  (TPAMI)}, 40(4):834--848, 2017.

\bibitem{cheng2021fashion}
Wen-Huang Cheng, Sijie Song, Chieh-Yun Chen, Shintami~Chusnul Hidayati, and
  Jiaying Liu.
\newblock Fashion meets computer vision: A survey.
\newblock {\em ACM Computing Surveys (CSUR)}, 54(4):1--41, 2021.

\bibitem{deng2019arcface}
Jiankang Deng, Jia Guo, Niannan Xue, and Stefanos Zafeiriou.
\newblock {ArcFace: Additive angular margin loss for deep face recognition}.
\newblock In {\em Computer Vision and Pattern Recognition (CVPR)}, pages
  4690--4699, 2019.

\bibitem{dong2019towards}
Haoye Dong, Xiaodan Liang, Xiaohui Shen, Bochao Wang, Hanjiang Lai, Jia Zhu,
  Zhiting Hu, and Jian Yin.
\newblock Towards multi-pose guided virtual try-on network.
\newblock In {\em International Conference on Computer Vision (ICCV)}, pages
  9026--9035, 2019.

\bibitem{fele2022c}
Benjamin Fele, Ajda Lampe, Peter Peer, and Vitomir Struc.
\newblock {C-VTON: Context-driven image-based virtual try-on network}.
\newblock In {\em Winter Conference on Applications of Computer Vision (WACV)},
  pages 3144--3153, 2022.

\bibitem{feng_virtual_2018}
Zhuming Feng, Fei Jiang, and Ruimin Shen.
\newblock Virtual glasses try-on based on large pose estimation.
\newblock 131:226--233.

\bibitem{ge2021parser}
Yuying Ge, Yibing Song, Ruimao Zhang, Chongjian Ge, Wei Liu, and Ping Luo.
\newblock Parser-free virtual try-on via distilling appearance flows.
\newblock In {\em Computer Vision and Pattern Recognition (CVPR)}, pages
  8485--8493, 2021.

\bibitem{gong2021aesthetics}
Wei Gong and Laila Khalid.
\newblock Aesthetics, personalization and recommendation: A survey on deep
  learning in fashion.
\newblock {\em arXiv preprint arXiv:2101.08301}, 2021.

\bibitem{goodfellow_generative_2014}
Ian Goodfellow, Jean Pouget-Abadie, Mehdi Mirza, Bing Xu, David Warde-Farley,
  Sherjil Ozair, Aaron Courville, and Yoshua Bengio.
\newblock Generative adversarial nets.
\newblock In {\em Advances in Neural Information Processing Systems (NeurIPS)},
  2014.

\bibitem{han2018viton}
Xintong Han, Zuxuan Wu, Zhe Wu, Ruichi Yu, and Larry~S Davis.
\newblock Viton: An image-based virtual try-on network.
\newblock In {\em Computer Vision and Pattern Recognition (CVPR)}, pages
  7543--7552, 2018.

\bibitem{harkonen2020ganspace}
Erik H{\"a}rk{\"o}nen, Aaron Hertzmann, Jaakko Lehtinen, and Sylvain Paris.
\newblock {GANspace: Discovering interpretable GAN controls}.
\newblock {\em Advances in Neural Information Processing Systems (NeurIPS)},
  33:9841--9850, 2020.

\bibitem{he2019attgan}
Zhenliang He, Wangmeng Zuo, Meina Kan, Shiguang Shan, and Xilin Chen.
\newblock Attgan: Facial attribute editing by only changing what you want.
\newblock {\em IEEE Transactions on Image Processing (TIP)}, 28(11):5464--5478,
  2019.

\bibitem{heusel2017gans}
Martin Heusel, Hubert Ramsauer, Thomas Unterthiner, Bernhard Nessler, and Sepp
  Hochreiter.
\newblock Gans trained by a two time-scale update rule converge to a local nash
  equilibrium.
\newblock {\em Advances in Neural Information Processing Systems (NeurIPS)},
  30, 2017.

\bibitem{huang_human-centric_2012}
Szu-Hao Huang, Yu-I Yang, and Chih-Hsing Chu.
\newblock Human-centric design personalization of 3d glasses frame in
  markerless augmented reality.
\newblock {\em Advanced Engineering Informatics}, 26(1):35--45, 2022.

\bibitem{harkonen_ganspace_2020}
Erik Härkönen, Aaron Hertzmann, Jaakko Lehtinen, and Sylvain Paris.
\newblock {GANSpace}: Discovering interpretable {GAN} controls.
\newblock In {\em Advances in Neural Information Processing Systems (NeurIPS)},
  volume~33, pages 9841--9850, 2020.

\bibitem{isola2017image}
Phillip Isola, Jun-Yan Zhu, Tinghui Zhou, and Alexei~A Efros.
\newblock Image-to-image translation with conditional adversarial networks.
\newblock In {\em Computer Vision and Pattern Recognition (CVPR)}, pages
  1125--1134, 2017.

\bibitem{issenhuth2020not}
Thibaut Issenhuth, J{\'e}r{\'e}mie Mary, and Cl{\'e}ment Calauz{\`e}nes.
\newblock Do not mask what you do not need to mask: a parser-free virtual
  try-on.
\newblock In {\em European Conference on Computer Vision (ECCV)}, pages
  619--635. Springer, 2020.

\bibitem{jiang2022clothformer}
Jianbin Jiang, Tan Wang, He Yan, and Junhui Liu.
\newblock Clothformer: Taming video virtual try-on in all module.
\newblock In {\em Computer Vision and Pattern Recognition (CVPR)}, pages
  10799--10808, 2022.

\bibitem{jong2020virtual}
Andrew Jong, Melody Moh, and Teng-Sheng Moh.
\newblock Virtual try-on with generative adversarial networks: A taxonomical
  survey.
\newblock In {\em Advancements in Computer Vision Applications in Intelligent
  Systems and Multimedia Technologies}, pages 76--100. IGI Global, 2020.

\bibitem{karras2018progressive}
Tero Karras, Timo Aila, Samuli Laine, and Jaakko Lehtinen.
\newblock Progressive growing of gans for improved quality, stability, and
  variation.
\newblock In {\em International Conference on Learning Representations (ICLR)},
  2018.

\bibitem{karras2020training}
Tero Karras, Miika Aittala, Janne Hellsten, Samuli Laine, Jaakko Lehtinen, and
  Timo Aila.
\newblock Training generative adversarial networks with limited data.
\newblock {\em Advances in Neural Information Processing Systems (NeurIPS)},
  33:12104--12114, 2020.

\bibitem{karras2021alias}
Tero Karras, Miika Aittala, Samuli Laine, Erik H{\"a}rk{\"o}nen, Janne
  Hellsten, Jaakko Lehtinen, and Timo Aila.
\newblock Alias-free generative adversarial networks.
\newblock {\em Advances in Neural Information Processing Systems (NeurIPS)},
  34:852--863, 2021.

\bibitem{karras2019style}
Tero Karras, Samuli Laine, and Timo Aila.
\newblock A style-based generator architecture for generative adversarial
  networks.
\newblock In {\em Computer Vision and Pattern Recognition (CVPR)}, pages
  4401--4410, 2019.

\bibitem{karras2020analyzing}
Tero Karras, Samuli Laine, Miika Aittala, Janne Hellsten, Jaakko Lehtinen, and
  Timo Aila.
\newblock Analyzing and improving the image quality of stylegan.
\newblock In {\em Computer Vision and Pattern Recognition (CVPR)}, pages
  8110--8119, 2020.

\bibitem{khodadadeh2022latent}
Siavash Khodadadeh, Shabnam Ghadar, Saeid Motiian, Wei-An Lin, Ladislau
  B{\"o}l{\"o}ni, and Ratheesh Kalarot.
\newblock Latent to latent: A learned mapper for identity preserving editing of
  multiple face attributes in stylegan-generated images.
\newblock In {\em Winter Conference on Applications of Computer Vision (WACV)},
  pages 3184--3192, 2022.

\bibitem{kwon2021diagonal}
Gihyun Kwon and Jong~Chul Ye.
\newblock Diagonal attention and style-based gan for content-style
  disentanglement in image generation and translation.
\newblock In {\em International Conference on Computer Vision (ICCV)}, pages
  13980--13989, 2021.

\bibitem{lee_maskgan_2020}
Cheng-Han Lee, Ziwei Liu, Lingyun Wu, and Ping Luo.
\newblock {MaskGAN}: Towards diverse and interactive facial image manipulation.
\newblock In {\em Computer Vision and Pattern Recognition ({CVPR})}, pages
  5548--5557, 2020.

\bibitem{lee2020maskgan}
Cheng-Han Lee, Ziwei Liu, Lingyun Wu, and Ping Luo.
\newblock {MaskGAN: Towards diverse and interactive facial image manipulation}.
\newblock In {\em Computer Vision and Pattern Recognition (CVPR)}, pages
  5549--5558, 2020.

\bibitem{liu2022towards}
Kanglin Liu, Gaofeng Cao, Fei Zhou, Bozhi Liu, Jiang Duan, and Guoping Qiu.
\newblock Towards disentangling latent space for unsupervised semantic face
  editing.
\newblock {\em IEEE Transactions on Image Processing (TIP)}, 31:1475--1489,
  2022.

\bibitem{marelli_faithful_2021}
Davide Marelli, Simone Bianco, and Gianluigi Ciocca.
\newblock Faithful fit, markerless, 3d eyeglasses virtual try-on.
\newblock In Alberto Del~Bimbo, Rita Cucchiara, Stan Sclaroff, Giovanni~Maria
  Farinella, Tao Mei, Marco Bertini, Hugo~Jair Escalante, and Roberto Vezzani,
  editors, {\em International Conference on Pattern Recognition {ICPR}:
  Workshops and Challenges}, pages 460--471, 2021.

\bibitem{niswar_virtual_2011}
Arthur Niswar, Ishtiaq~Rasool Khan, and Farzam Farbiz.
\newblock Virtual try-on of eyeglasses using 3d model of the head.
\newblock In {\em 10th International Conference on Virtual Reality Continuum
  and Its Applications in Industry (VRCAI)}, pages 435--438, 2011.

\bibitem{parmar2022spatially}
Gaurav Parmar, Yijun Li, Jingwan Lu, Richard Zhang, Jun-Yan Zhu, and
  Krishna~Kumar Singh.
\newblock Spatially-adaptive multilayer selection for gan inversion and
  editing.
\newblock In {\em Computer Vision and Pattern Recognition (CVPR)}, pages
  11399--11409, 2022.

\bibitem{pernuvs2021high}
Martin Pernu{\v{s}}, Vitomir {\v{S}}truc, and Simon Dobri{\v{s}}ek.
\newblock {High Resolution Face Editing with Masked GAN Latent Code
  Optimization}.
\newblock {\em IEEE Transactions on Image Processing (TIP), MR}, 2022.

\bibitem{pizzati2021comogan}
Fabio Pizzati, Pietro Cerri, and Raoul de Charette.
\newblock Comogan: continuous model-guided image-to-image translation.
\newblock In {\em Computer Vision and Pattern Recognition (CVPR)}, pages
  14288--14298, 2021.

\bibitem{press_emerging_2018}
Ori Press, Tomer Galanti, Sagie Benaim, and Lior Wolf.
\newblock Emerging disentanglement in auto-encoder based unsupervised image
  content transfer.
\newblock In {\em International Conference on Learning Representations (ICLR)},
  2018.

\bibitem{remy2016style}
Nathalie Remy, Eveline Speelman, and Steven Swartz.
\newblock Style that’s sustainable: A new fast-fashion formula.
\newblock Technical report, McKinsey Global Institute, 2016.

\bibitem{shen2020interpreting}
Yujun Shen, Jinjin Gu, Xiaoou Tang, and Bolei Zhou.
\newblock Interpreting the latent space of gans for semantic face editing.
\newblock In {\em Computer Vision and Pattern Recognition (CVPR)}, pages
  9243--9252, 2020.

\bibitem{shen2021interfacegan}
Yujun Shen, Ceyuan Yang, Xiaoou Tang, and Bolei Zhou.
\newblock {InterfaceGAN: Interpreting the disentangled face representation
  learned by GANs}.
\newblock {\em IEEE Transactions on Pattern Analysis and Machine Intelligence
  (TPAMI)}, 2022.

\bibitem{tolosana2020deepfakes}
Ruben Tolosana, Ruben Vera-Rodriguez, Julian Fierrez, Aythami Morales, and
  Javier Ortega-Garcia.
\newblock Deepfakes and beyond: A survey of face manipulation and fake
  detection.
\newblock {\em Information Fusion}, 64:131--148, 2020.

\bibitem{tov2021designing}
Omer Tov, Yuval Alaluf, Yotam Nitzan, Or Patashnik, and Daniel Cohen-Or.
\newblock {Designing an encoder for StyleGAN image manipulation}.
\newblock {\em ACM Transactions on Graphics (TOG)}, 40(4):1--14, 2021.

\bibitem{Vieira2014}
Tiago~F. Vieira, Andrea Bottino, Aldo Laurentini, and Matteo De~Simone.
\newblock Detecting siblings in image pairs.
\newblock {\em The Visual Computer}, 30(12):1333--1345, 2014.

\bibitem{wang2022high}
Tengfei Wang, Yong Zhang, Yanbo Fan, Jue Wang, and Qifeng Chen.
\newblock High-fidelity gan inversion for image attribute editing.
\newblock In {\em Computer Vision and Pattern Recognition (CVPR)}, pages
  11379--11388, 2022.

\bibitem{wang2018high}
Ting-Chun Wang, Ming-Yu Liu, Jun-Yan Zhu, Andrew Tao, Jan Kautz, and Bryan
  Catanzaro.
\newblock High-resolution image synthesis and semantic manipulation with
  conditional gans.
\newblock In {\em Computer Vision and Pattern Recognition (CVPR)}, pages
  8798--8807, 2018.

\bibitem{wang2021generative}
Zhengwei Wang, Qi She, and Tomas~E Ward.
\newblock {Generative adversarial networks in computer vision: A survey and
  taxonomy}.
\newblock {\em ACM Computing Surveys (CSUR)}, 54(2):1--38, 2021.

\bibitem{xia2022gan}
Weihao Xia, Yulun Zhang, Yujiu Yang, Jing-Hao Xue, Bolei Zhou, and Ming-Hsuan
  Yang.
\newblock {GAN inversion: A survey}.
\newblock {\em IEEE Transactions on Pattern Analysis and Machine Intelligence
  (TPAMI)}, 2022.

\bibitem{xu2022transeditor}
Yanbo Xu, Yueqin Yin, Liming Jiang, Qianyi Wu, Chengyao Zheng, Chen~Change Loy,
  Bo Dai, and Wayne Wu.
\newblock Transeditor: Transformer-based dual-space gan for highly controllable
  facial editing.
\newblock In {\em Computer Vision and Pattern Recognition (CVPR)}, pages
  7683--7692, 2022.

\bibitem{yang2021l2m}
Guoxing Yang, Nanyi Fei, Mingyu Ding, Guangzhen Liu, Zhiwu Lu, and Tao Xiang.
\newblock L2m-gan: Learning to manipulate latent space semantics for facial
  attribute editing.
\newblock In {\em Computer Vision and Pattern Recognition (CVPR)}, pages
  2951--2960, 2021.

\bibitem{yang2020towards}
Han Yang, Ruimao Zhang, Xiaobao Guo, Wei Liu, Wangmeng Zuo, and Ping Luo.
\newblock Towards photo-realistic virtual try-on by adaptively
  generating-preserving image content.
\newblock In {\em Computer Vision and Pattern Recognition (CVPR)}, pages
  7850--7859, 2020.

\bibitem{yuan_magic_2017}
Xiaoyun Yuan, Difei Tang, Yebin Liu, Qing Ling, and Lu Fang.
\newblock Magic glasses: From 2d to 3d.
\newblock {\em {IEEE} Transactions on Circuits and Systems for Video Technology
  (TCSVT)}, 27(4):843--854, 2017.

\bibitem{zhang_augmented_2018}
Boping Zhang.
\newblock Augmented reality virtual glasses try-on technology based on {iOS}
  platform.
\newblock {\em {EURASIP} Journal on Image and Video Processing}, 2018.1:1--19,
  2018.

\bibitem{zhang_virtual_2017}
Qian Zhang, Yu Guo, Pierre-Yves Laffont, Tobias Martin, and Markus Gross.
\newblock A virtual try-on system for prescription eyeglasses.
\newblock {\em {IEEE} Computer Graphics and Applications}, 37(4):84--93, 2017.

\bibitem{zhang2021datasetgan}
Yuxuan Zhang, Huan Ling, Jun Gao, Kangxue Yin, Jean-Francois Lafleche, Adela
  Barriuso, Antonio Torralba, and Sanja Fidler.
\newblock {DatasetGAN}: Efficient labeled data factory with minimal human
  effort.
\newblock In {\em Computer Vision and Pattern Recognition (CVPR)}, pages
  10145--10155, 2021.

\bibitem{zhu2017unpaired}
Jun-Yan Zhu, Taesung Park, Phillip Isola, and Alexei~A Efros.
\newblock Unpaired image-to-image translation using cycle-consistent
  adversarial networks.
\newblock In {\em International Conference on Computer Vision (ICCV)}, pages
  2223--2232, 2017.

\end{thebibliography}
}

\clearpage

\appendix

\begin{Large}
  \noindent \textbf{GlassesGAN: Supplementary Material}\\ \\%
\end{Large}

In the main part of the paper, we presented a wide range of results to highlight the merits of the proposed GlassesGAN framework for personalization of glasses in  virtual try-on settings. In this \textit{Supplementary material}, we now show additional results to further highlight the capabilities of GlassesGAN. Specifically, we: $(i)$ demonstrate the robustness of the overall processing pipeline on another dataset (MetFaces) with out-of-domain images, $(ii)$ show additional continuous multi-style editing results, $(iii)$ provide additional details on the user-study conducted, $(iv)$ describe ablation experiments with respect to the Targeted Subspace Modelling (TSM), and $(v)$ provide information on the reproducibility of our results.

\section{Out of Domain Edits} \label{appendix:add datasets}

\begin{figure}[t]
\begin{center}
\includegraphics[width=1.0\columnwidth]{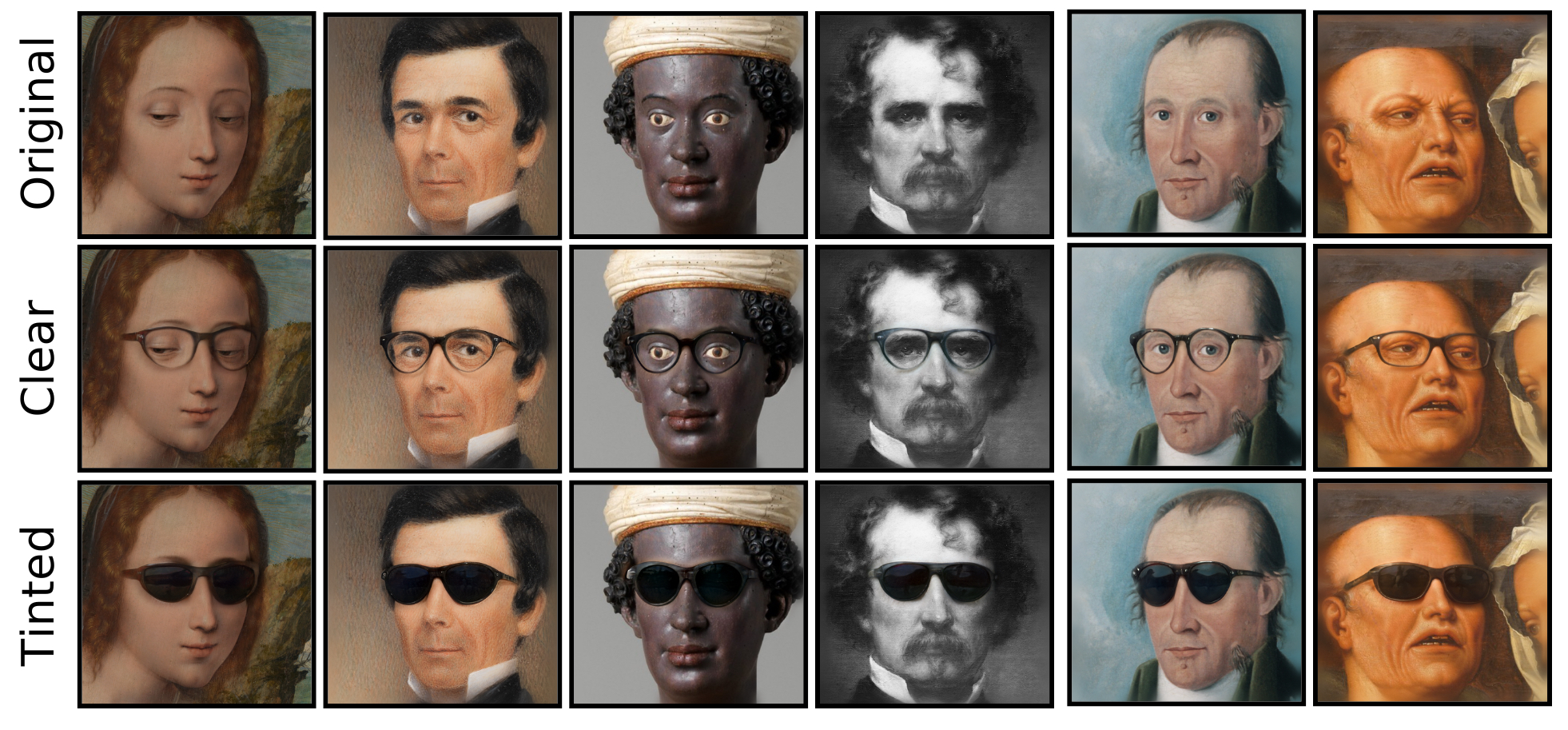}
    \caption{\textbf{Out of domain editing with GlassesGAN.} Additions of clear and tinted glasses to images from the MetFaces dataset are presented. Observe the realism and detail of the generated edits.}\vspace{-4mm}
  \label{fig:out-of-domain-edit}
\end{center}
\end{figure}

To demonstrate the robustness of the processing pipeline of GlassesGAN, we generate a few example edits on the MetFaces dataset. MetFaces~\cite{karras2020training} is a dataset of human faces extracted from works of art. Because the images in this dataset do not correspond to real faces, they come with vastly different characteristics than the facial images used to train our framework. As can be seen from Figure~\ref{fig:out-of-domain-edit}, GlassesGAN is able to apply clear and tinted glasses to the
artwork without any changes to the overall framework. We
observe that even with these challenging images, the edits
appear highly realistic and visually convincing.

\section{Additional Visual Results} \label{appendix:add vis results}

\begin{figure*}[t]
\begin{center}
  \includegraphics[width=0.95\linewidth]{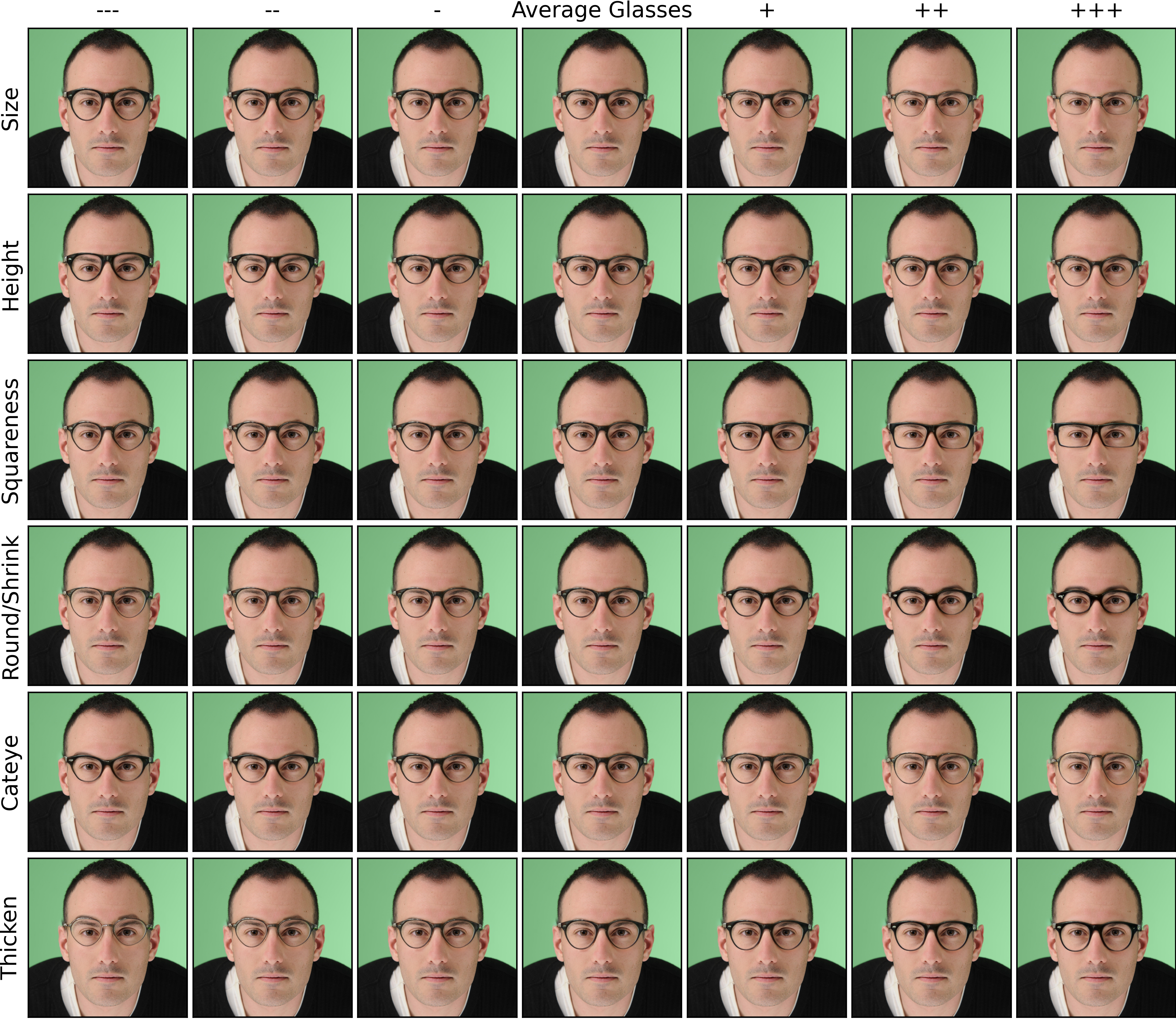}\vspace{-2mm}
    \caption{Extended visualization of continuous multi-style edits with clear glasses on sample from SiblingsDB-HQf.}
  \label{fig:contmultiedit__DSC5792}
\end{center}
\end{figure*}

\begin{figure*}[t]
\begin{center}
  \includegraphics[width=1.0\linewidth]{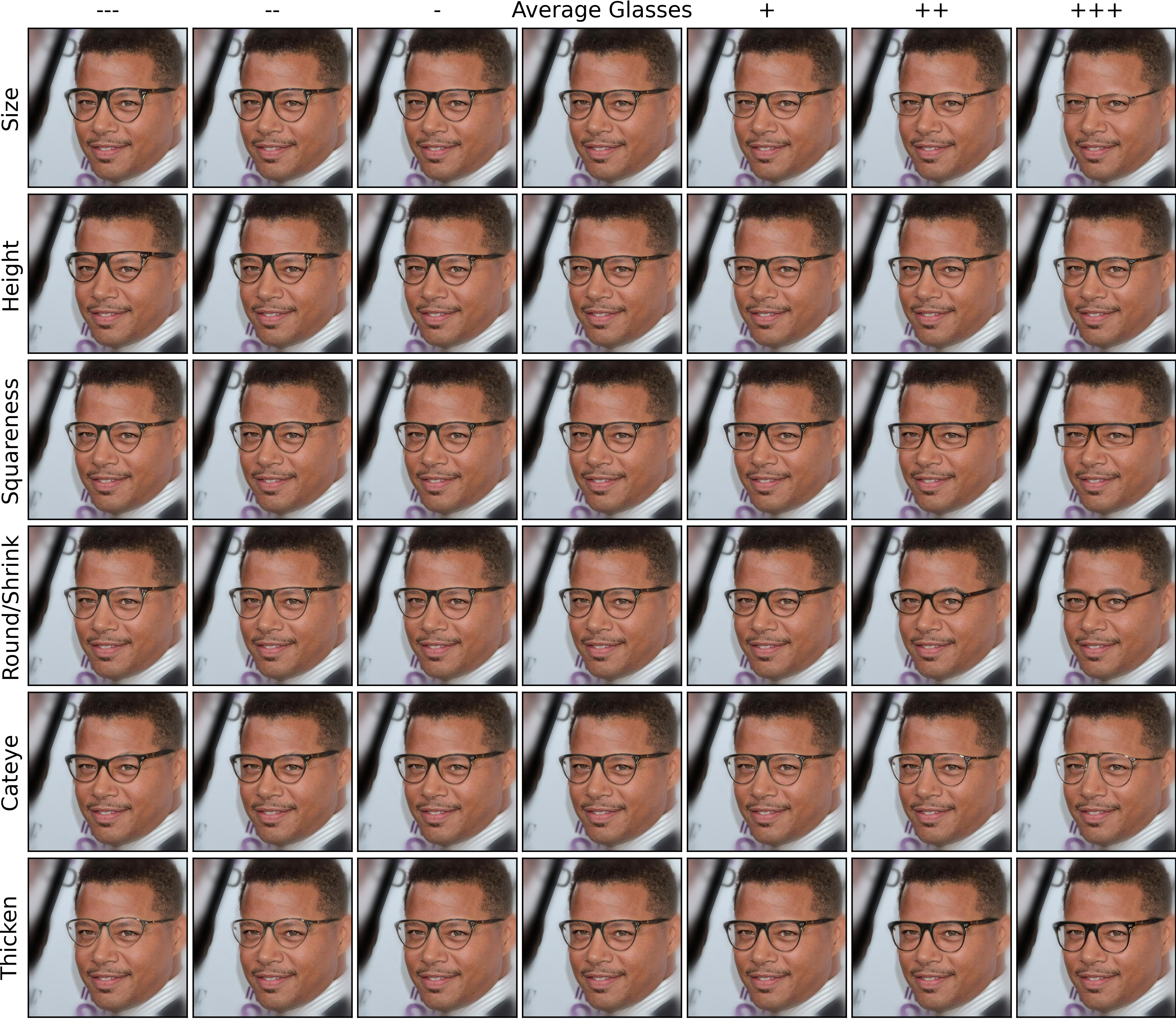}\vspace{-2mm}
    \caption{Extended visualization of continuous multi-style edits with clear glasses on sample from CelebA-HQ.}
  \label{fig:contmultiedit_29974}
\end{center}
\end{figure*}

\begin{figure*}[t]
\begin{center}
  \includegraphics[width=1.0\linewidth]{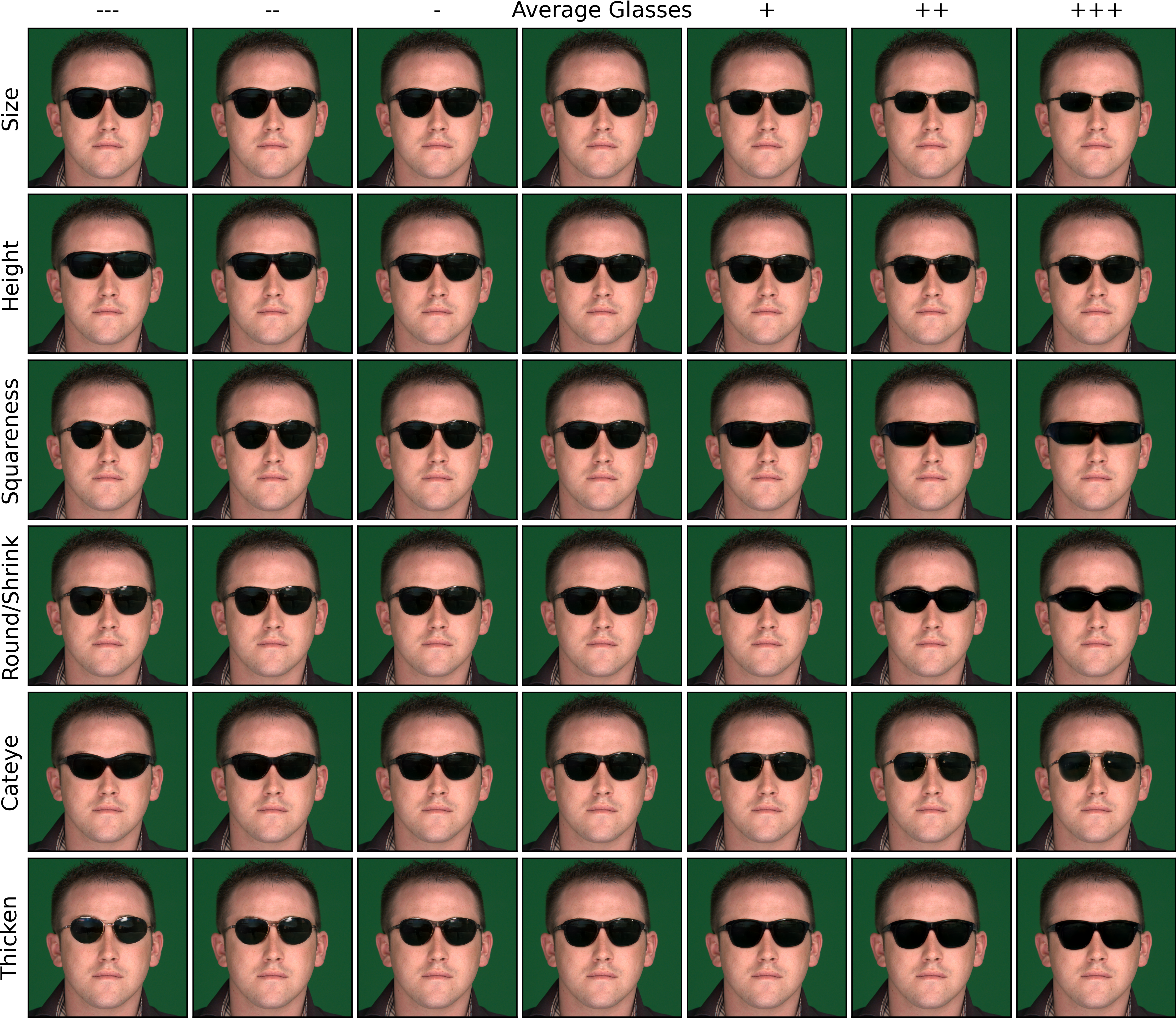}\vspace{-2mm}
    \caption{Extended visualization of continuous multi-style edits with tinted glasses on sample from SiblingsDB-HQf.}
  \label{fig:contmultiedit__DSC6005}
\end{center}
\end{figure*}

\begin{figure*}[t]
\begin{center}
  \includegraphics[width=1.0\linewidth]{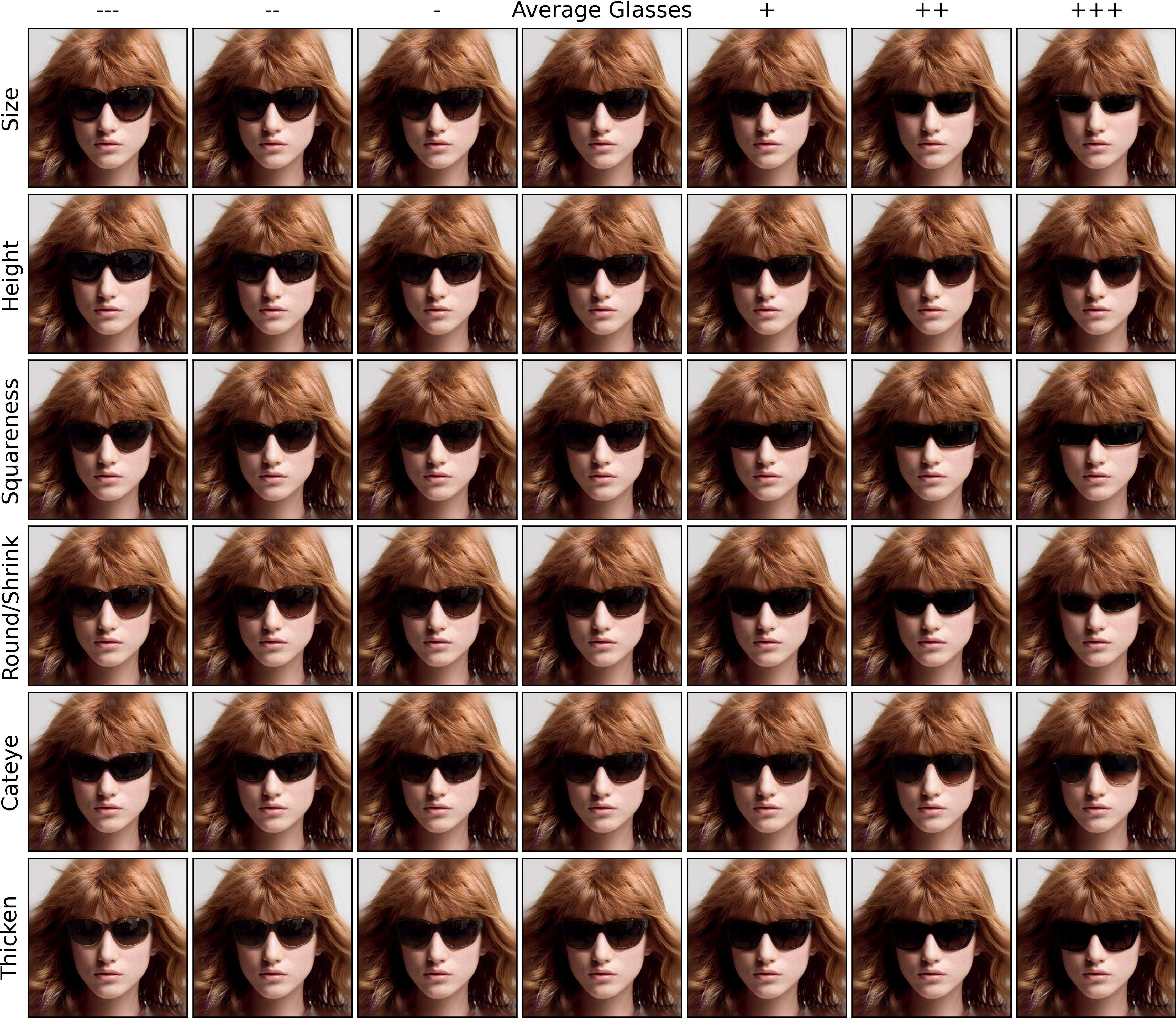}\vspace{-2mm}
    \caption{Extended visualization of continuous multi-style edits with tinted glasses on sample from CelebA-HQ.}
  \label{fig:contmultiedit_29981}
\end{center}
\end{figure*}

In Figures~\ref{fig:contmultiedit__DSC5792} and \ref{fig:contmultiedit_29974} we show additional high-resolution results of continuous multi-style edits with a couple of sample images and clear glasses. Similarly, in Figures~\ref{fig:contmultiedit__DSC6005} and \ref{fig:contmultiedit_29981} we demonstrate continuous multi-style edits for tinted glasses. In order to better demonstrate the continuous edit-capability of GlassesGAN, we also provide a video demonstration of our editing as part of the supplementary material. We observe that GlassesGAN is able to smoothly transition between different eyeglass styles in the latent space while preserving the realism of the edit.

\section{Additional User Survey Details}

\begin{figure*}[t]
\begin{center}
\includegraphics[width=1.00\linewidth, trim = 46cm 17.8cm 0cm 0cm,  clip]{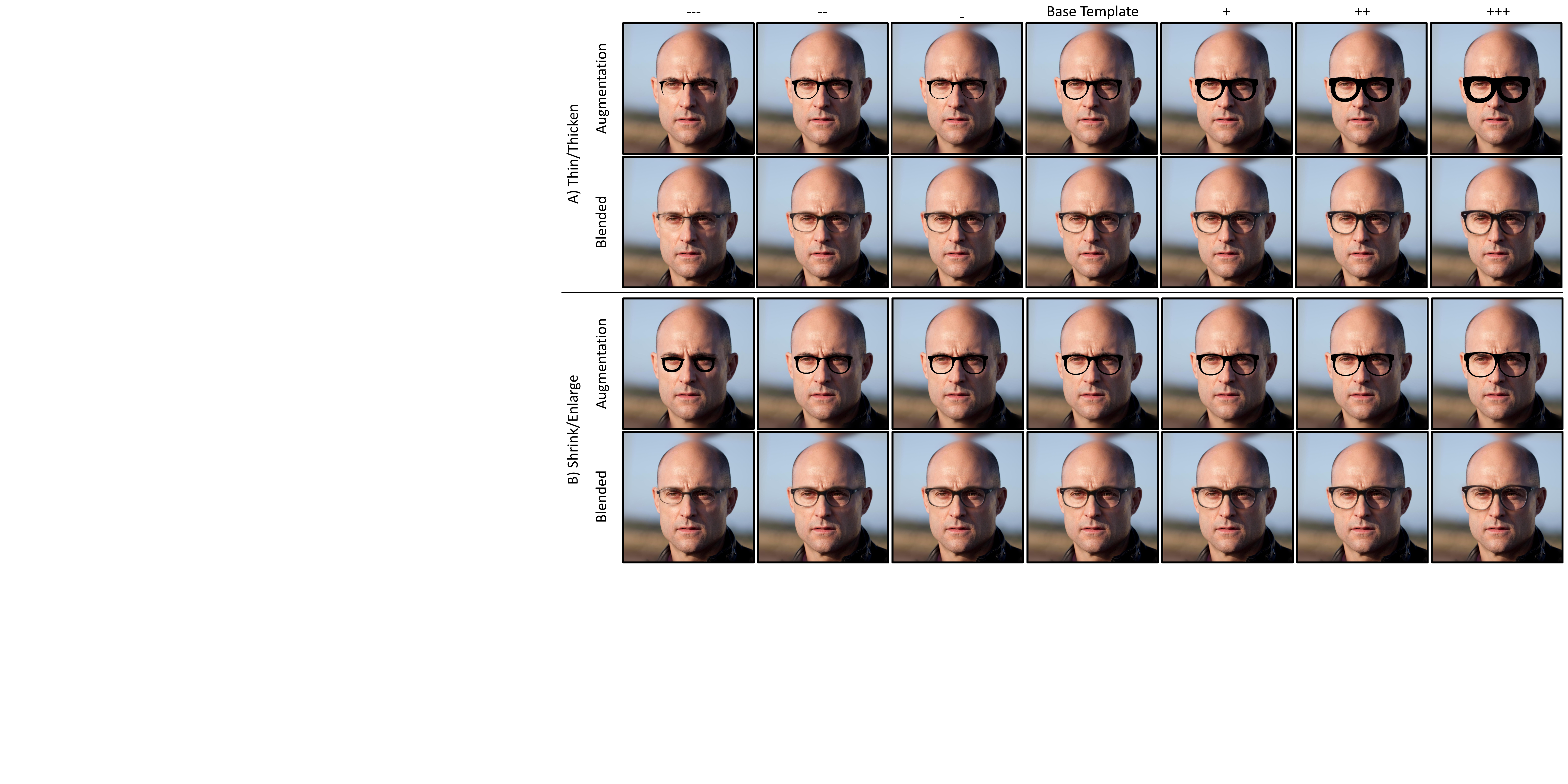} %
    \caption{\textbf{Targeted Subspace Modeling (TSM) Ablation Results Demonstrating Edit Flexibility Without TSM.} Without TSM, glasses personalization via continuous edits is substantially limited. In this figure, we show that our template augmentation procedure is only able to (A) thin and thicken or (B) shrink and enlarge glasses frames, as compared to the six distinct edit styles available when using TSM. The upper row in each group shows the augmented image while the lower row shows the result after encoding and blending.}
  \label{fig:edit_flexibility}
\end{center}
\end{figure*}

\begin{figure*}[t]
\begin{center}
\includegraphics[width=0.50\linewidth, trim = 15cm 27cm 0cm 0cm,  clip]{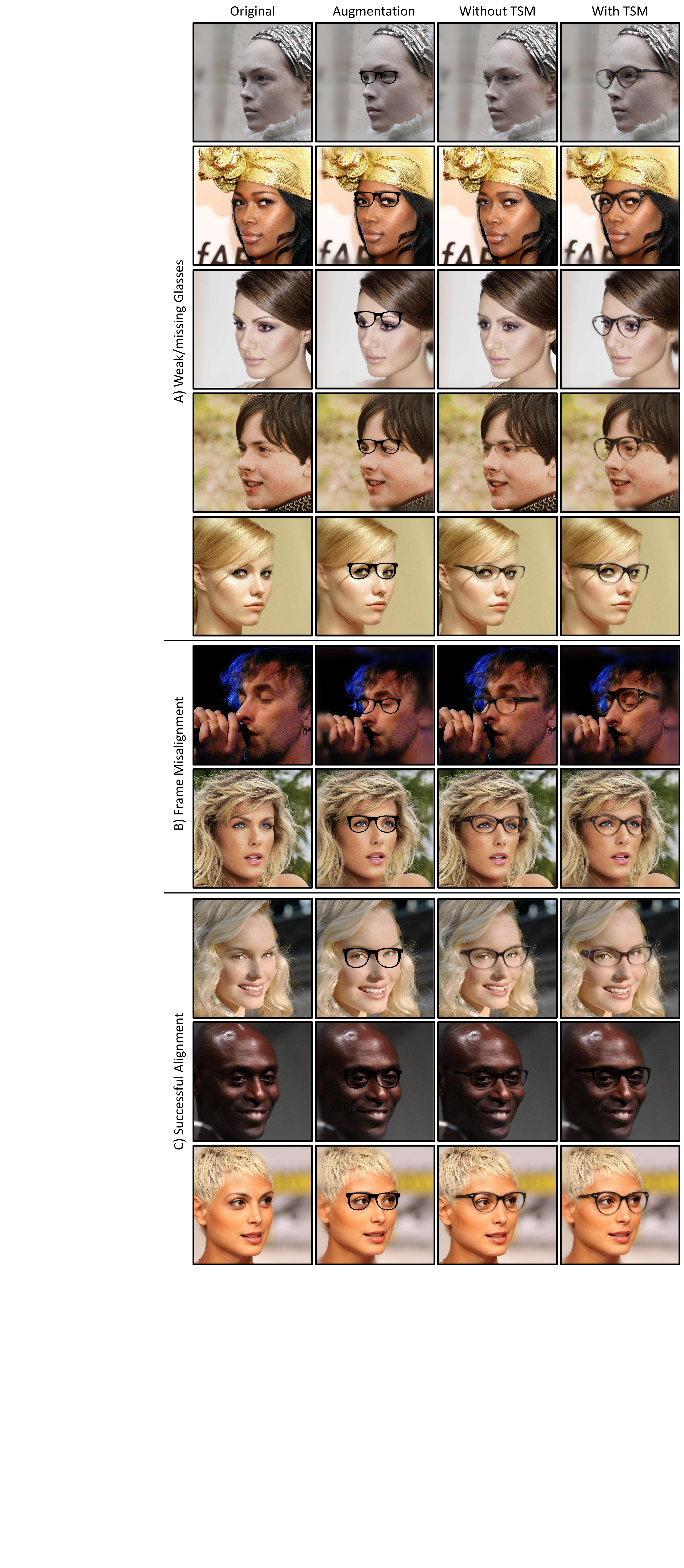} %
    \caption{\textbf{Targeted Subspace Modeling (TSM) Ablation Results for Off-angle Samples.} The first column shows the original probe images while the second shows the intermediate augmentation image from the eyeglasses masking. The third and forth columns show the final edited image without and with TSM respectively. Group A and B show samples where removing TSM caused glasses to be poorly embedded and misoriented respectively. The samples in group C show the result when the augmentation alignment is correct.}
  \label{fig:blend_back_baseline}
\end{center}
\end{figure*}

Evaluators were shown randomly selected probe images and randomly ordered glasses edit images from each comparison method and asked to choose the edit image that best meets the criteria of the prompt question. Prompt questions were $Q1$: Which of the edited images preserves the identity of the original image the most?, $Q2$: Which of the edited images is of the highest quality overall?, $Q3$: Which of the edited images has the most realistic looking and visually convincing glasses? (consider realism, shadows, frame shape, fit on ears, etc.), and $Q4$: Which of the edited images is closest to what you would consider a good try on result? %

\section{TSM Ablation Experiment}

Targeted Subspace Modeling (TSM), introduced in this paper, is a key component of GlassesGAN that allows us to capture the simulated variations in the appearance of eyeglasses in a small number of principal subspace axes. In this section, we investigate the importance of Targeted Subspace Modeling (TSM) in the methodology of GlassesGAN. To do this, we explore an alternate procedure where the desired personalization to eyeglasses is performed to the augmentation masks rather than in the learned glasses subspace of the generator's latent space, thereby bypassing TSM. Instead of learning the glasses subspace, this alternative method creates the desired glasses shape in the augmentation mask, sequentially encodes and then decodes the probe image with the mask applied, and blends the result into the original image using the face parser $S$. We found that this alternate methodology is able to provide a virtual-try-on experience, but suffers when it comes to flexibility of edits, off-angle robustness, and inference-time computational complexity. We elaborate on these characteristics in the following sections. 

\subsection{Edit Flexibility}
As described in Section~\ref{sec:SAD}, the Synthetic Appearance Discovery (SAD) mechanism begins with a set of hand-drawn eyeglass templates and then augments them using morphological operations, such as dilation and erosion. With only these augmented templates available, the alternative methodology without TSM is substantially limited in the type of continuous edits available. In Figure \ref{fig:edit_flexibility}, we show our ability to thin and thicken frames (\ref{fig:edit_flexibility}A) and shrink and enlarge frames (\ref{fig:edit_flexibility}B) without TSM.
With use of TSM, on the other hand, GlassesGAN is able to learn six distinct edit types including the rounding and squaring of glasses and cat-eye appearances as shown in Figures \ref{fig:contmultiedit__DSC5792}, \ref{fig:contmultiedit_29974}, \ref{fig:contmultiedit__DSC6005}, \ref{fig:contmultiedit_29981}. 

\subsection{Off-angle Robustness}
The removal of TSM from the methodology also comes at the expense of the robustness of glasses edits to off-angle scenarios, as shown in Figure \ref{fig:blend_back_baseline}. As the eyeglasses positioning and size are based on the coordinates of the temples, there is no mechanism to warp the mask to fit the face correctly in severely off-angle poses. This causes eyeglasses to be poorly embedded (\ref{fig:blend_back_baseline}A), misoriented (\ref{fig:blend_back_baseline}B), or correctly oriented when the augmentation positioning is close enough (\ref{fig:blend_back_baseline}C).

\subsection{Inference Computation}

During inference-time sequential edits without TSM are more time-intensive. This is because every edit without TSM requires a run from the Encoder model $E$. Alternatively, the use of TSM moves the editing step into the latent space, thereby only requiring encoding to occur during the first run. Since GlassesGAN is designed for continuous edits to eyeglass style, this extra latency for sequential edits can quickly become very apparent.

\section{Reproducibility}

All of our experiments are fully reproducible. The models used for the implementation of GlassesGAN are all publicly available from the official repositories, i.e.: %
\begin{itemize}
    \item StyleGAN2: \\
    {\small \url{https://github.com/NVlabs/stylegan2}} 
    \item E4E encoder: \\
    {\small \url{https://github.com/omertov/encoder4editing}}
    \item DatasetGAN: \\
    {\small \url{https://github.com/nv-tlabs/datasetGAN_release}}
    \item DeepLabV2: \\
    {\small \url{https://github.com/tensorflow/models/tree/master/research/deeplab}}
    \item Dlib: \\
    {\small \url{http://dlib.net/}}
    \item ArcFace: \\
    {\small \url{https://github.com/deepinsight/insightface}}
\end{itemize}

Additionally, we also plan to publicly release the GlassesGAN source code, including all training and testing scripts, once the review procedure is completed.

\end{document}